\documentclass[runningheads]{llncs}

\usepackage{lmodern}
\usepackage{eccv}

\definecolor{zoey}{rgb}{0.12, 0.47, 0.71}

\usepackage{multirow} 
\usepackage{colortbl}
\usepackage{xcolor}
\usepackage{booktabs}
\definecolor{headerblue}{RGB}{180, 195, 210}
\definecolor{darow}{RGB}{255, 235, 210}
\definecolor{groupgray}{RGB}{235, 235, 235}
\usepackage{eccvabbrv}

\usepackage{graphicx}
\usepackage{booktabs}

\usepackage[accsupp]{axessibility}  

\usepackage{hyperref}

\usepackage{orcidlink}
\usepackage{marvosym} 

\begin{document}


\title{Lost in the Tail: Addressing Geographic Imbalance in Urban Visual Place Recognition}

\titlerunning{Lost in the Tail: Geographic Imbalance in Urban VPR}

\author{Zhiyao Shu\inst{1}\orcidlink{0009-0001-9423-0207} \and
Jiacheng Yang\inst{2}\orcidlink{0009-0006-5224-3217} \and
Yang Lu\inst{2}\orcidlink{0000-0002-3497-9611} \and
Waishan Qiu\inst{3}\orcidlink{0000-0001-6461-7243} \and
Chuan Li\inst{4} \and
Da Chen\inst{5}\orcidlink{0000-0002-5062-0270}\textsuperscript{\Letter}}

\authorrunning{Z.Shu et al.}

\institute{
$^1$\,George Mason University, Fairfax, USA \quad
$^2$\,Xiamen University, Xiamen, China \quad
$^3$\,University of Hong Kong, Hong Kong, China\\
$^4$\,Lambda, San Francisco, USA \quad
$^5$\,University of Bath, Bath, UK\\
\email{zshu2@gmu.edu, \{jiachengyang,luyang\}@xmu.edu.cn, waishanq@hku.hk, c@lambdal.com, da.chen@bath.edu}
}

\makeatletter{\renewcommand{\thefootnote}{}\let\Hy@raisedlink\@gobble\footnotetext{\Letter~Corresponding author: Da Chen. \email{da.chen@bath.edu}}}\makeatother

\maketitle

\begin{abstract}
\noindent Urban-scale Visual Place Recognition (VPR) aims to identify the geographic location of a query image by matching it against a geo-tagged database. While recent methods achieve impressive performance, they overlook a serious long-tailed problem hidden in urban-scale datasets, which biases the model towards locations with abundant images and ignores less-visited areas, causing models to systematically favor frequently photographed locations while failing in sparsely covered areas. In this paper, we systematically characterize this imbalance challenge and propose \textbf{Distribution-Aware Place Recognition (DAPR)}, a model-agnostic plug-in framework that rebalances gradient contributions across head and tail classes. 
Additionally, within classification-retrieval pipelines, DAPR applies a multi-scale distance search mechanism to compute per-class distributional compactness, providing complementary gains at the retrieval stage. On the large-scale SF-XL benchmark, our framework outperforms the previous classification-retrieval baseline by \textbf{18.3\%} on test set v1, and \textbf{6.7\%} on test set v2. As a plug-in module, it achieves consistent improvements across representative VPR methods on SF-XL, MSLS, and Pitts30k, demonstrating broad generalizability across different methods and benchmarks. 



\end{abstract}
\section{Introduction}
\label{sec:intro}

\begin{figure}[t]
  \centering
  \includegraphics[width=0.91\linewidth]{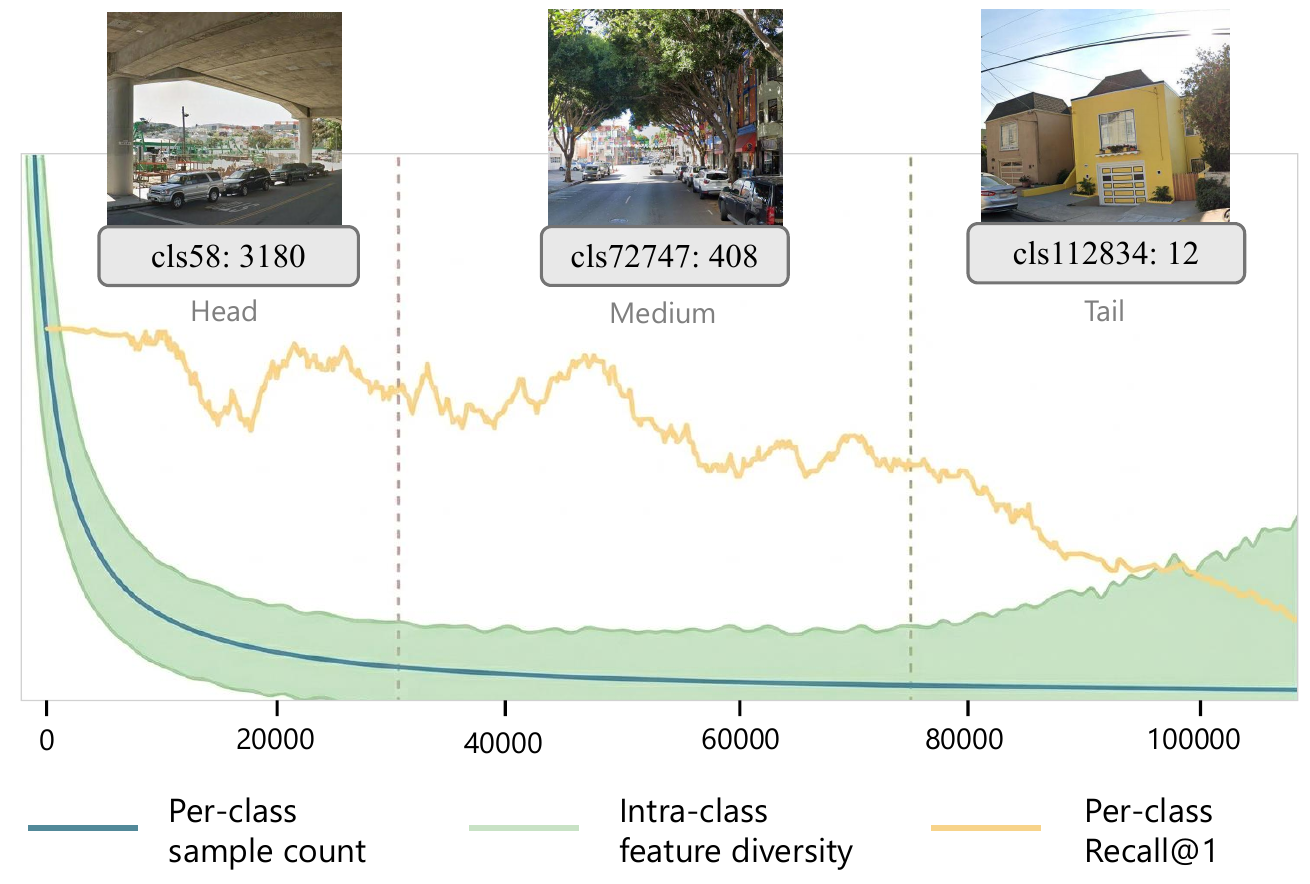}
  \caption{Geographic classes in SF-XL~\cite{berton2022rethinking} are ranked by sample count (blue curve) and partitioned into head (top 30\%), middle (40\%), and tail (bottom 30\%). Intra-class feature diversity (green band) indicates richer and more varied visual content in sparsely sampled regions (widens toward tail classes). Per-class Recall@1 (orange curve), measured on the D\&C~\cite{Trivigno_2023_divideclassify} model, declines sharply toward the tail.}
   \vspace{-3mm}
   \label{fig:teaser}
\end{figure}

Visual Place Recognition (VPR), also known as geo-localization, is a fundamental task in Computer Vision that aims to determine the geographic location of a query image by matching it against a geo-tagged reference database. 
Recent proposed VPR methods learn global descriptors for direct similarity search, ranging from lightweight CNN-based approaches~\cite{ali2023mixvpr, barbarani2023local, Arandjelovic_2016_CVPR} to transformer-based aggregation methods~\cite{Izquierdo_2024_salad, ali2024boq, lu2024cricavpr, zhang2025efficient}. 
Alternatively, some methods frame VPR as a classification problem, partitioning the map into discrete spatial cells to learn discriminative place descriptors~\cite{berton2022rethinking, berton2023eigenplaces, seo2018cplanet}, with D\&C~\cite{Trivigno_2023_divideclassify} further introducing a classification-retrieval pipeline (hereafter denoted as \textit{mixed-pipeline}) to improve inference efficiency. These VPR approaches achieve strong accuracy on existing small-scale and large-scale benchmarks.

However, existing methods overlooked the severe long-tailed distribution challenge in the existing geographic datasets. Current VPR datasets are sourced from Google Street View~\cite{berton2022rethinking, Arandjelovic_2016_CVPR, ali2022gsv} or crowd-sourced platforms~\cite{warburg2020mapillary}, where image density is governed by vehicle traffic volume and photographer frequency rather than geographic feature richness. Consequently, environments that are visually distinctive and rich in diverse geographic features, such as residential streets, alleys and low-traffic urban corridors, accumulate only sparse training coverage.
Figure~\ref{fig:teaser} illustrates the extent of this imbalance on SF-XL~\cite{berton2022rethinking}. We rank all geographic classes by image count and partition them into head (top 30\% by frequency), middle (40\%), and tail (bottom 30\%), exposing a $\sim$300:1 imbalance ratio where head classes contain over 3{,}000 images while tail classes have as few as~12. Notably, intra-class feature diversity (green band) widens toward the tail, reflecting greater variance in pairwise feature similarity within sparsely sampled regions. This indicates that tail-class areas contain richer and more varied visual content. Per-class Recall@1 (orange curve, measured on D\&C~\cite{Trivigno_2023_divideclassify}) declines sharply toward the tail, revealing a compounding failure: the classes that are inherently hardest to recognize due to their visual diversity receive the least training supervision.

This limitation fundamentally undermines the reliability of deployed real-world VPR systems. For example, urban safety applications such as autonomous patrol robots~\cite{ye2024human} are specifically deployed in low-traffic, GPS-degraded zones such as the residential and peripheral urban areas that constitute tail classes in existing datasets. A VPR model that has never adequately learned the distinctive features of these regions will suffer the worst localization failures at the locations of highest operational demand, leading to safety-critical risks where reliable spatial awareness is most needed.

\begin{figure}[t]
  \centering
   \includegraphics[width=0.95\linewidth]{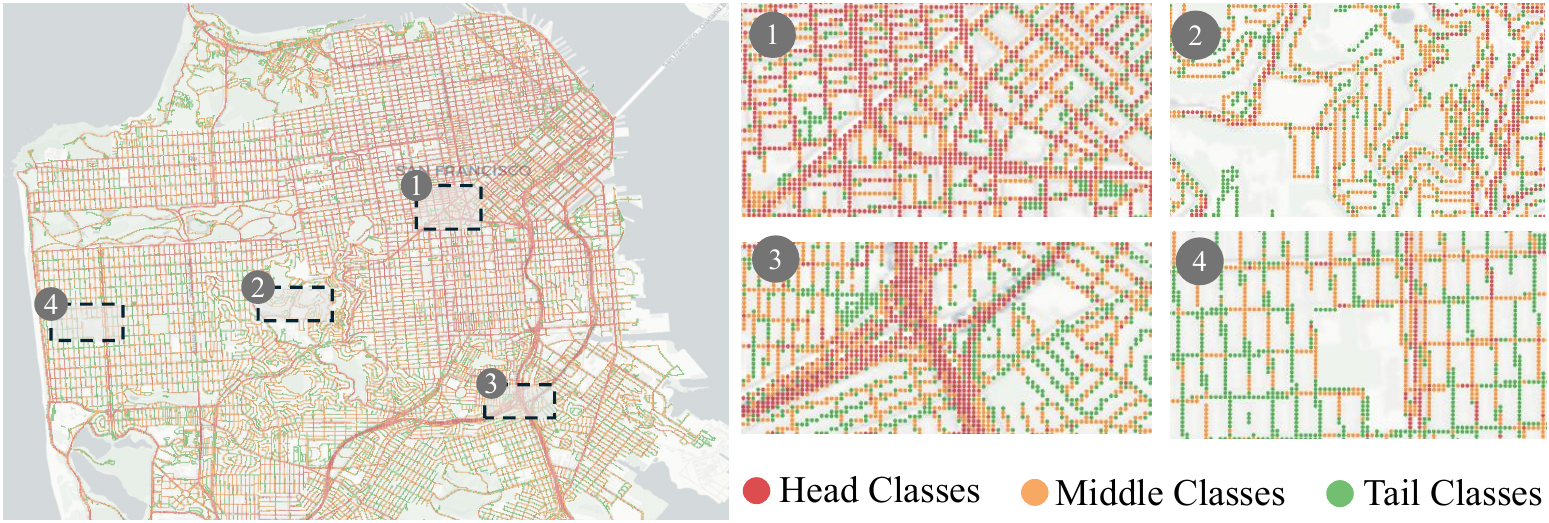}
   \caption{The geographic classes distribution of SF\_XL~\cite{berton2022rethinking} on real-world map. Red regions (head classes) represent frequently photographed locations from SF\_XL~\cite{berton2022rethinking}. Orange regions (middle classes) show moderately sampled areas, and green regions (tail classes) indicate sparsely photographed locations. Four representative regions (1-4) are zoom-in spatial patterns of class imbalance: region 1 and 4 reveal predominantly head classes in major roads, regions 2-3 demonstrate mixed distributions. 
   }
   \vspace{-3mm}
   \label{fig:class_spatial_distribution}
\end{figure}

Furthermore, direct similarity search over millions of images on a large-scale benchmark, such as SF-XL~\cite{berton2022rethinking} containing 41.2 million images, introduces non-trivial computational overhead at inference.
While mixed-pipeline methods~\cite{Trivigno_2023_divideclassify} address this computational challenge by filtering candidate classes before similarity search and enabling practical deployment on large-scale datasets, their reliance on L2 distance during inference treats all feature distributions uniformly, implicitly assuming equal feature compactness across geographic classes. This causes head classes to dominate matching due to their dense and compact representations, while tail-class features are systematically underweighted despite their discriminative value.

To address these challenges, we propose a Distribution-Aware Place Recognition (DAPR) framework, which balances the feature distribution with a distribution-aware loss strategy (named as Low-visit Bias (LB) loss $\mathcal{L}_{lb}$) and a multi-scale distance search mechanism. In the learning phase, DAPR rebalances gradient contributions across classes through inverse-frequency weighting and corrects classifier bias via prior-based logit adjustment. Within mixed-pipelines, the multi-scale distance search mechanism introduces Characteristic Function~\cite{feuerverger1977empirical} to adaptively weight amplitude (distributional compactness) and phase (directional similarity) components of feature distances, enabling fair comparison between head and tail classes without requiring explicit labels at inference time.
We summarize the contributions as follows:

\begin{itemize}
    \item We identify and formalize the long-tailed geographic distribution problem in urban-scale VPR, revealing that class imbalance with spatial bias fundamentally limits existing VPR methods.

    \item We propose \textbf{DAPR}, a model-agnostic plug-in framework that addresses long-tailed geographic imbalance across different VPR methods. Within mixed pipelines, we further introduce a Multi-scale Distance Search Mechanism to provide complementary distribution-aware gains at the retrieval stage.
    
    \item Comprehensive experiments on SF-XL~\cite{berton2022rethinking} demonstrate that DAPR achieves state-of-the-art performance, surpassing the previous best mixed-pipeline baseline by \textbf{+18.3\%} R@1 on test~v1 and \textbf{+6.7\%} on test~v2, while being over 60$\times$ faster than full-database retrieval methods. As a plug-in module, $\mathcal{L}_{lb}$ yields consistent improvements when integrated into representative retrieval methods (SALAD, BoQ) across MSLS and Pitts30k, confirming that long-tailed geographic imbalance is a universal bottleneck in urban-scale VPR.   
\end{itemize}
\section{Related Work}

\subsection{Visual Place Recognition}
\label{subsec:vpr_related}

VPR research addresses localization at different spatial scales~\cite{zhu2021vigor, hu2018cvm,wang2022transvpr,zhang2023etr, paolicelli2022learning,Izquierdo_2024_salad, lu2024towards, keetha2023anyloc, izquierdo2024close,Trivigno_2023_divideclassify}. At the global scale, systems recognize landmarks and determine geographic coordinates across vast areas by leveraging highly distinctive features within regional-level classification~\cite{haas2024pigeon}. Beyond spatial scale, cross-view matching introduces additional complexity, where query and reference images are captured from fundamentally different viewpoints, requiring viewpoint-invariant representations~\cite{zhu2021vigor, hu2018cvm}.
Urban-scale VPR focuses on fine-grained localization within city environments~\cite{ge2020self, berton2022deep}, where the primary challenge lies in retrieving visually similar images based on robust feature matching. Recent feature extraction methods have shifted from CNN-based architectures~\cite{Arandjelovic_2016_CVPR, radenovic2018fine} to transformer-based models leveraging self-attention~\cite{wang2022transvpr}, hybrid-attention~\cite{hu2024enhancing}, and multi-attention mechanisms~\cite{zhang2023etr, paolicelli2022learning}. In particular, DINOv2 has demonstrated strong capabilities for extracting semantically rich features that generalize well across diverse urban environments~\cite{Izquierdo_2024_salad, lu2024towards, keetha2023anyloc, izquierdo2024close}.

Beyond feature extraction, aggregation strategies further improve descriptor quality~\cite{babenko2015aggregating}. VLAD-based methods~\cite{jegou2010aggregating, khaliq2024vlad_buff} pool local features into compact global representations, while attentional pyramid pooling approaches~\cite{peng2021attentional} enhance descriptor quality through multi-scale spatial encoding. 
Two-stage retrieval pipelines have become prevalent~\cite{lu2024towards, zhu2023r2former}, first retrieving coarse candidates via global descriptors, then applying geometric verification or re-ranking with local features.
Single-stage methods rely solely on global features, with recent DINOv2-based 
approaches such as SALAD~\cite{Izquierdo_2024_salad}, BoQ~\cite{ali2024boq}, CricaVPR~\cite{lu2024cricavpr}, and EffoVPR~\cite{tzachor2024effovpr} establishing strong baselines across benchmarks~\cite{shao2023global, garg2024revisit}. CliqueMining~\cite{izquierdo2024close} mines visually similar place cliques to sharpen geographic distance sensitivity, implicit aggregation~\cite{lu2026towards} aggregates features within the transformer backbone itself. And MegaLoc~\cite{berton2025megaloc} unifies training across datasets for strong cross-domain generalization.
Methods that treat VPR as a geographic classification problem partition the map into discrete geo-cells and train a classifier over these place categories using large-margin cosine losses~\cite{wang2018cosface} for discriminative descriptor learning~\cite{berton2022rethinking,
berton2023eigenplaces, seo2018cplanet}.
At inference, D\&C~\cite{Trivigno_2023_divideclassify} extend this trained classifier to filter the full database down to a small candidate pool, after which standard L2-based retrieval re-ranks the candidates.

\begin{figure*}[h]

  \centering
    \includegraphics[width=\linewidth]{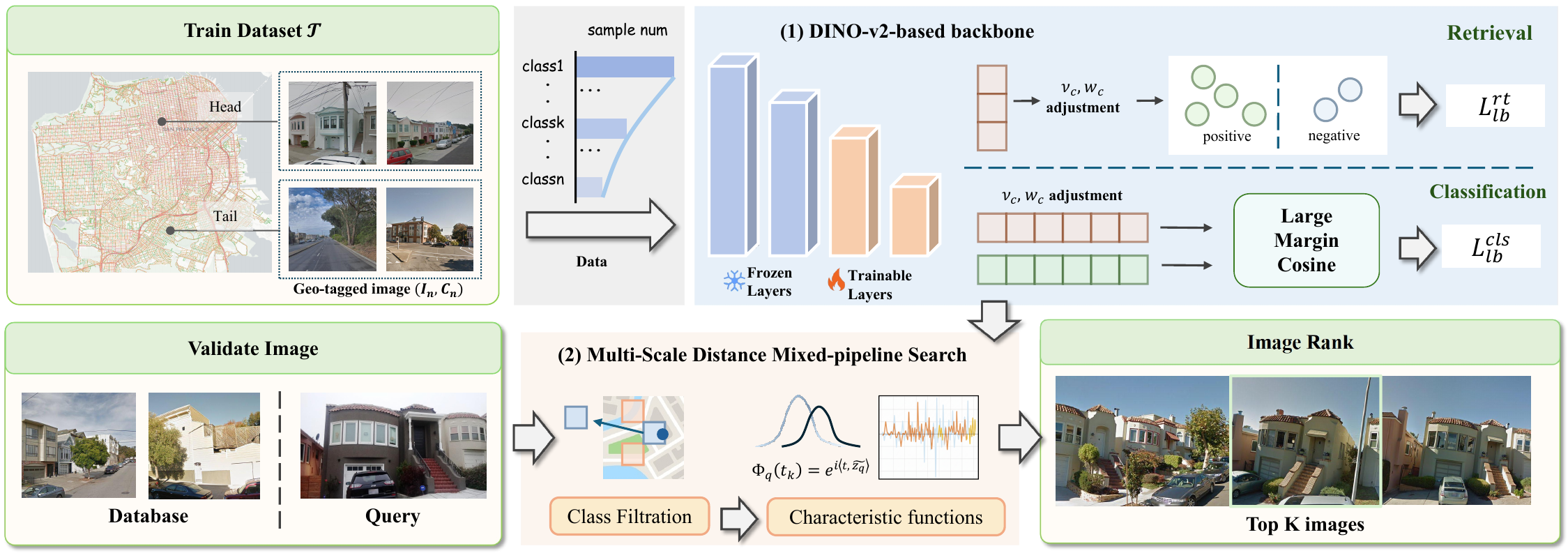}
    \caption{\textbf{Distribution-Aware Place Recognition Framework.} The geo-tagged images are processed through (1) A DINOv2 backbone extracts features that are optimized under two paradigms: retrieval methods with $\mathcal{L}_{lb}^{rt}$, while classification methods with Large Margin Cosine producing $\mathcal{L}_{lb}^{cls}$. Both apply class-dependent weights $w_c$ and logit adjustment $\nu_c$ to rebalance gradient contributions across head and tail classes. (2) Mixed-pipeline filters the database to candidate classes, which are then re-ranked using Characteristic Function Distance (CFD) in the frequency domain to return the top-$K$ matched images.}
    \vspace{-3mm}
    \label{fig:framework}
\end{figure*}

\vspace{-2mm}
\subsection{Long-Tailed Distribution for Imbalanced Data}
\label{subsec:LT_related}

Real-world data are inherently imbalanced—a small number of head classes account for the majority of samples, whereas the more numerous tail classes contain only a few instances ~\cite{andersonLongTailWhy2006,yu2025reviving, wang_kill_2024, shi_how_2023, chen2021self}. This long-tailed distribution biases empirical-risk-minimization models toward the head and leads to degraded performance on the tail~\cite{chu2020feature,wang2023unified, yang2024harnessing,wang2025unified,wu2020distribution}. Existing long-tailed learning methods can be broadly grouped into three families: data resampling~ \cite{yu2025reviving, wang_kill_2024}, training framework modifications ~\cite{shi_how_2023, chu2020feature}, and loss-function designs ~\cite{wu2020distribution}. Among these, loss-based methods are widely adopted for downstream tasks because they require no extra training stages and are plug-and-play.
For instance, ~\cite{laloss2020} proposed the logit adjustment (LA) loss, which explicitly calibrates class confidences according to class frequency to mitigate head bias. ~\cite{wu2020distribution} introduced the Distribution-Balanced (DB) loss, combining rebalanced weighting with negative-tolerant regularization to alleviate the suppression of tail classes by negatives. ~\cite{zhao2022adaptive} incorporated the notion of “hard classes,” assigning larger weights to harder-to-learn categories. Moreover, ~\cite{kangdecoupling} decoupled representation learning from classification and employed a cosine classifier to reduce the impact of imbalance. Building on these insights, in this paper, we design a tail-favoring loss, i.e., Low-visit Bias (LB) loss $\mathcal{L}_{lb}$, based on class frequency for VPR, which not only achieves better performance on tail classes, but also improves the performance of the head and middle classes.
\vspace{-0.5em}
\section{Method}
\vspace{-0.5em}

As illustrated in Fig.~\ref{fig:framework}, the proposed DAPR framework consists of two plug-in modules. 
First, the Low-visit Bias (LB) loss addresses long-tailed geographic imbalance by correcting gradient contributions across head and tail 
classes (\S\ref{sec:daloss}). Second, for the mixed-pipeline where the classifier filters the database to a candidate pool during inference, we propose a multi-scale distance search (\S\ref{sec:cfd}) that operates in the frequency domain to provide multi-scale, distribution-aware similarity for retrieval within the filtered candidates.

\vspace{-2mm}
\subsection{Problem Formulation}
Let $\mathcal{D} = \{(\mathbf{I}_i, \mathbf{p}_i)\}_{i=1}^{N_\mathcal{D}}$ denote a geo-tagged reference database, where $\mathbf{I}_i \in \mathbb{R}^{H \times W \times 3}$ is the $i$-th image and $\mathbf{p}_i = (x_i, y_i) \in \mathbb{R}^2$ its spatial coordinates. Given a query image $\mathbf{I}_q$ at an unknown location $\mathbf{p}_q$, VPR aims to retrieve the reference image(s) from $\mathcal{D}$ geographically closest to $\mathbf{p}_q$. A large-scale training set $\mathcal{T} = \{(\mathbf{I}_n, \mathbf{p}_n)\}_{n=1}^{N_\mathcal{T}}$ is used to learn a feature extractor $f\colon \mathbb{R}^{H \times W \times 3} \!\to\! \mathbb{R}^d$~\cite{Arandjelovic_2016_CVPR, ge2020self, jin2017learned, liu2019stochastic}.

Following the classification-based paradigm~\cite{berton2022rethinking, berton2023eigenplaces, Trivigno_2023_divideclassify}, which avoids the convergence difficulty of direct ranking optimization on large-scale datasets~\cite{Arandjelovic_2016_CVPR, radenovic2018fine}, we partition the geographic space into $C$ square grid cells of side length $M$ meters:
\begin{equation}
\mathcal{C}_{i,j} = \{(x, y) \in \mathbb{R}^2 : \lfloor x/M \rfloor = i,\; \lfloor y/M \rfloor = j\}.
\end{equation}
Each image $\mathbf{I}_n$ is assigned a class label $C_n$ from its grid cell, yielding class frequencies $\hat{p}_c = n_c / N_\mathcal{T}$, where $n_c$ denotes the number of samples in class $c$. At inference, the classifier filters the database to a candidate pool $\mathcal{D}_q \subset \mathcal{D}$ of predicted classes, within which similarity-based retrieval is performed. However, real-world urban datasets exhibit severely long-tailed class distributions, causing both training bias and retrieval degradation for under-represented tail classes.

\vspace{-2mm}
\subsection{Geographic Long-Tail Aware Training}
\label{sec:daloss}
To address the systematic under-representation of low-frequency geographic locations, we propose the \textbf{Low-visit Bias (LB) loss ($\mathcal{L}_{lb}$)}, a plug-and-play training module that corrects the systematic bias caused by long-tailed geographic distributions.
As discussed, the long-tailed issue hidden in VPR datasets biases the model toward head classes, leading to degraded performance on tail classes. Specifically, this bias is mainly caused by the disproportionately larger gradients contributed by frequent areas during optimization~\cite{li2022long,andersonLongTailWhy2006}. For this reason, we integrate three coupled mechanisms: class-distribution reweighting, prior-based logit adjustment, and large-margin cosine similarity, instantiated as $\mathcal{L}_{lb}^{cls}$ for classification (Eq.~\ref{eq:db_loss}) and $\mathcal{L}_{lb}^{rt}$ for retrieval (Eq.~\ref{eq:rt_loss}).
First, to rebalance gradient contributions across head and tail areas, each class is assigned a weight $w_c$ inversely proportional to its frequency $\hat{p}_c = n_c / N_\mathcal{T}$, where $n_c$ is the number of samples in class $c$ and $N$ is the total number of samples:

\begin{equation}
w_c = \frac{C}{\sum_{c'=1}^C (\hat{p}_{c'} + \epsilon)^{-\beta}} \cdot (\hat{p}_c + \epsilon)^{-\beta},
\end{equation}

\noindent where $\beta \in [0 , 1]$ controls the strength of reweighting, and $\epsilon$ helps avoid numerical instability. This weighting mechanism effectively amplifies the contribution of tail classes and prevents gradient explosion, while the normalization term ensures stable gradient magnitudes throughout training.

Second, to further mitigate bias caused by class prior encoding in decision boundaries, a prior-based logit adjustment is applied as:
\begin{equation}
\nu_c = -\kappa \cdot \log(\frac{\hat{p}_c}{1 - \hat{p}_c}),
\end{equation}

\noindent where $\kappa \in [0, 1]$ controls the correction strength. 
Since $\nu_c>0$ for tail classes, their logits are upweighted, promoting class-agnostic representation learning and reducing head-class bias.

Finally, after obtaining weighting and logit adjustment parameters, the Low-visit Bias loss function $\mathcal{L}_{lb}$ is designed to improve both retrieval and classification tasks. For the classification task, Large Margin Cosine Similarity $d$~\cite{wang2018cosface} is first computed to normalize both feature embeddings and classifier weights to the unit hypersphere as follows.

\begin{equation}
    d = \frac{\mathbf{z}^\top W_j}{\|\mathbf{z}\| \|W_j\|} \\,
\end{equation}
\begin{equation}
    z_j = s \cdot (d - m \cdot \mathbb{I}[j = c]),
\end{equation}
where $\mathbf{z}$ is the feature embedding for training sample $(\mathbf{I}_n, C_n)$, and $W_j$ is the $j$-th classifier weight vector. $s$ is a scale factor, $m$ is the angular margin, and $\mathbb{I}[j = c]$ is the indicator function. This enforces minimum angular separation between classes in the embedding space. By doing this, the model can capture clear inter-class boundaries and learn compact, discriminative features across all classes. Then, $w_c$ and $\nu_c$ are employed to enable robust and balanced representation learning across the long-tailed geographic distribution inherent in real-world VPR datasets by:

\begin{equation}
\mathcal{L}_{lb}^{cls} = -w_c \cdot \log\left(\frac{\exp(z_c - \nu_c)}{\sum_{j=1}^{C} \exp(z_j - \nu_j)}\right).
\label{eq:db_loss}
\end{equation}

This function simultaneously addresses gradient imbalance through sample weighting $w_c$, removes classifier bias through logit adjustment $\nu_c$, and enforces feature discriminability through the angular margin embedded in $z_j$.

For the retrieval-based task, feature similarity $d_{ij}^{f}$ is first computed for sample $i$ and $j$ by $s_{ij} = \mathbf{f}_i^\top \mathbf{f}_j$, where $\mathbf{f}$ is the extracted feature, then combining widely used Multi-Similarity, the LB loss can be written as:

\begin{equation}
\begin{aligned}
\label{eq:rt_loss}
\mathcal{L}_{lb}^{rt}
=&
\frac{1}{N}
\sum_{i=1}^{N}
w_{y_i}
\Bigg[
\frac{1}{\gamma_p}
\log
\left(
1
+
\sum_{\substack{j \neq i \\ y_j = y_i}}
\exp
\left(
-\gamma_p ( d_{ij}^{f} - \tau )
\right)
\right)
\\
&+
\frac{1}{\gamma_n}
\log
\left(
1
+
\sum_{\substack{j \\ y_j \neq y_i}}
\exp
\left(
\gamma_n
\left(
d_{ij}^{f}
-
\nu_{y_j}
-
\tau
\right)
\right)
\right)
\Bigg],
\end{aligned}
\end{equation}
where $N$ denotes the mini-batch size,
$\gamma_p$ and $\gamma_n$ are the positive and negative scaling factors, respectively, and $\tau$ is a similarity threshold. The index set 
$\{ j \mid j \neq i,\, y_j = y_i \}$ 
corresponds to positive samples of anchor $i$, while $\{ j \mid y_j \neq y_i \}$ 
denotes the set of negative samples. By doing this, the retrieval model can learn a more robust and balanced feature space for VPR datasets. 

\vspace{-2mm}
\subsection{Multi-scale Distance Search Mechanism}
\label{sec:cfd}
While $\mathcal{L}_{lb}$ mitigates long-tail bias during training, distributional imbalance persists at retrieval time, particularly on large-scale datasets such as SF-XL. Within the classifier-filtered candidate pool $\mathcal{D}_q$, tail-class features remain sparse and follow complex, non-Gaussian distributions~\cite{zhang2023deep}, whereas head-class features form tight, numerically dominant clusters. A static metric such as $L_2$ or cosine similarity treats both regimes identically, causing queries from tail classes to systematically match nearby head-class clusters. 

To address this bias, we adopt the \textbf{Characteristic Function Distance (CFD)} as the retrieval metric, mapping features into the frequency domain~\cite{wang2025dataset, feuerverger1977empirical} for multi-scale similarity computation.

Let $\mathcal{S}_q$ denote the query feature set and $\mathcal{S}_j$ the descriptor set of candidate class $j$ within the filtered pool $\mathcal{D}_q$, where $\mathcal{S}_q$ reduces to the single query descriptor. By definition, their empirical characteristic functions~\cite{feuerverger1977empirical} are the sample averages over these sets, evaluated at $K$ frequency points $\mathbf{T} = \{\mathbf{t}_k\}_{k=1}^K$:

\begin{equation} \Phi_q(\mathbf{t}_k) = \frac{1}{|\mathcal{S}_q|}\sum_{\mathbf{z} \in \mathcal{S}_q} e^{i \langle \mathbf{t}_k, \mathbf{z} \rangle}, \quad \Phi_j(\mathbf{t}_k) = \frac{1}{|\mathcal{S}_j|}\sum_{\mathbf{z} \in \mathcal{S}_j} e^{i \langle \mathbf{t}_k, \mathbf{z} \rangle}, \end{equation}

\noindent where $i = \sqrt{-1}$, so that the modulus $|\Phi(\mathbf{t}_k)| \in [0, 1]$ is large when the set is compact and small when it is dispersed. To capture both fine-grained local patterns and global structure, we employ stratified multi-scale sampling for $\mathbf{T}$, drawing from Gaussian distributions at four logarithmically spaced scales. Please see the supplementary material for more details.

Each CF is then decomposed via Euler's formula into complementary components:

\begin{align}
\Phi(\mathbf{t}_k) = |\Phi(\mathbf{t}_k)| e^{i\alpha(\mathbf{t}_k)},
\end{align}

\noindent where $|\Phi(\mathbf{t}_k)|$ is the amplitude and $\alpha(\mathbf{t}_k) = \arg(\Phi(\mathbf{t}_k))$ is the phase. Head classes maintain more consistent amplitudes due to their dense feature clusters, while tail classes exhibit lower and more variable amplitudes, indicating dispersed, under-trained features.
For this reason, to enable distribution-aware matching, we compute mean amplitudes for query and database features:

\begin{equation}
\quad \bar{A}_q = \frac{1}{K}\sum_{k=1}^{K} |\Phi_q(\mathbf{t}_k)|, \quad \bar{A}_j = \frac{1}{K}\sum_{k=1}^{K} |\Phi_j(\mathbf{t}_k)|.
\end{equation}
The reliability ratio $r^{(j)} = \frac{\bar{A}_q}{\bar{A}_j}$ captures relative compactness, and the adaptive weights follow $\alpha_w^{(j)} = \min(\alpha \cdot r^{(j)}, 1), \lambda_w^{(j)} = 1 - \alpha_w^{(j)},$ with $\alpha \in [0, 1]$ a base amplitude weight.
At each frequency $\mathbf{t}_k$, the amplitude difference is:
\begin{equation}
D_{\text{amp}}^{(k)} = (|\Phi_q(\mathbf{t}_k)| - |\Phi_j(\mathbf{t}_k)|)^2.
\end{equation}

\noindent and the phase difference $D_{\text{phase}}^{(k)}$ using circular distance with wrapping at $2\pi$:
\begin{equation}
D_{\text{phase}}^{(k)} = \min(\delta^2, (2\pi - \delta)^2), \quad\delta = |\alpha_q(\mathbf{t}_k) - \alpha_j(\mathbf{t}_k)|.
\end{equation}

Finally, the CFD averages the component differences across all frequencies with adaptive weighting by:
\begin{equation}
D_{\text{CFD}}(q, j) = \alpha_w^{(j)} \cdot \frac{1}{K}\sum_{k=1}^{K} D_{\text{amp}}^{(k)} + \lambda_w^{(j)} \cdot \frac{1}{K}\sum_{k=1}^{K} D_{\text{phase}}^{(k)}.
\end{equation}
This distribution-aware adaptation enables fair similarity assessment without requiring explicit class labels at inference, and can be directly integrated with existing pre-trained VPR models. 
The amplitude ratio $r^{(j)}$ doubles as a per-query confidence estimate: a query whose descriptor distribution is compact relative to a candidate yields a high $r^{(j)}$ and a more reliable match. CFD thus reads an uncertainty signal from the descriptor distribution on the same retrieval pass, where existing methods obtain a comparable estimate by learning a stochastic embedding~\cite{warburg2021bayesian}, measuring the spread of the top-$K$ reference poses~\cite{zaffar2024sue}, or running a post-hoc local-feature matcher~\cite{sferrazza2025match}.
Note that CFD is applied only at the inference stage of the mixed pipeline since computing characteristic functions over all $C > 110{,}000$ class prototypes during training would scale as $\mathcal{O}(B \times C \times K \times D)$ per iteration, which is computationally prohibitive. 
We detail the uncertain experiment and CFD benefit to tail classes in the supplementary material.
\vspace{-0.5em}
\section{Experiments}
\vspace{-0.5mm}
We conduct experiments to evaluate DAPR from three perspectives. First, we compare DAPR against state-of-the-art VPR methods on the large-scale SF-XL benchmark (§\ref{sec:comparison}).
Second, we verify the generalization of $\mathcal{L}_{lb}$ by integrating it as a plug-in module into representative classification-based and retrieval-only methods on SF-XL, MSLS, and Pitts30k (§\ref{sec:generalization}). 
Third, we conduct ablation studies to isolate the individual contributions of $\mathcal{L}_{lb}$ and the CFD retrieval module, and benchmark them against alternative long-tailed learning strategies (§\ref{sec:ablation}).
\vspace{-2mm}
\subsection{Experimental Setup}
\noindent\textbf{Datasets.} We adopt the evaluation protocol from prior large-scale VPR works~\cite{Trivigno_2023_divideclassify, berton2022rethinking} and conduct experiments on four benchmarks. The large-scale SF-XL benchmark~\cite{berton2022rethinking} covers San Francisco with over 41M database images and two query sets (v1 and v2) captured under varied conditions. As shown in Figure~\ref{fig:teaser}, SF-XL exhibits a pronounced long-tail distribution. To demonstrate generalization beyond a single city, we evaluate the proposed module on MSLS~\cite{warburg2020mapillary}, covering diverse global urban environments, and Pitts30k~\cite{torii201524}, a standard medium-scale benchmark built at Pittsburgh. We further include Nordland~\cite{sunderhauf2013nordland}, to assess robustness under extreme seasonal appearance change. Long-tailed analysis for MSLS and Pitts30k is provided in the supplementary material.

\noindent\textbf{Implementation Details.}
All experiments are conducted on a single NVIDIA RTX 3090 GPU. For optimal performance, we employ DINO-v2~\cite{oquab2023dinov2} as the backbone, serving as the feature extractor. We train the model for 200 epochs using the Adam optimizer~\cite{adam2014method} with a batch size of 256 and a learning rate of $6 \times 10^{-6}$. More details are provided in the supplementary material.

\noindent\textbf{Evaluation Metrics.} In this paper, we adopt standard VPR evaluation protocols from prior work~\cite{Arandjelovic_2016_CVPR, berton2021viewpoint, ge2020self, hausler2021patch}. For classification-only methods~\cite{muller2018geolocation, seo2018cplanet}, we report Localization Radius@N (LR@N) using a 25-meter localization threshold, following~\cite{Trivigno_2023_divideclassify}. For measuring the search accuracy, we report Recall@N (R@N), defined as the percentage of queries where at least one of the top-N predictions falls within a 25-meter radius of the ground truth location. In fact, LR@N is equivalent to the R@N@25m used in retrieval according to~\cite{Trivigno_2023_divideclassify}. For simplicity, we use R@N for all evaluations in this paper.

\begin{table}[t]
    \centering
    \scriptsize
    \setlength{\tabcolsep}{4pt}
    \renewcommand{\arraystretch}{1.15}
    \begin{tabular}{l|c|c|c|c}
        \toprule
        \rule{0pt}{2.3ex}
        \textbf{Method} & \textbf{Backbone} & \textbf{\shortstack{Infer.\\Time}} & \textbf{R@1 v1} & \textbf{R@1 v2} \\
        \toprule
        \multicolumn{5}{l}{\textit{Classification}} \\
        PlaNet~\cite{weyand2016planet}                & EfficientNet  & 12 ms    & 24.5 & 53.1 \\
        HGE~\cite{muller2018geolocation}            & EfficientNet  & 15 ms    & 27.0 & 56.4 \\
        CPlaNet~\cite{seo2018cplanet}               & EfficientNet  & 17 ms    & 27.4 & 64.1 \\
        D\&C~\cite{Trivigno_2023_divideclassify}    & EfficientNet  & 12 ms    & 61.0 & 79.1 \\
        \hline
        DAPR - C                                    & EfficientNet & -- & 67.2 & 84.1\\
        DAPR - C                                    & ResNet101 & -- & 68.9 & 85.5 \\
        DAPR - C                                    & Dinov2 & -- & 86.4 & 91.1 \\
        \midrule
        \multicolumn{5}{l}{\textit{Retrieval}} \\
        NetVLAD~\cite{Arandjelovic_2016_CVPR}       & VGG16         & 12117 ms & 40.0 & 71.1 \\
        SFRS~\cite{ge2020self}                      & VGG16         & 12117 ms & 51.2 & 83.1 \\
        GeM~\cite{radenovic2018fine}                & VGG16         & 12117 ms & 21.7 & 43.1 \\
        CosPlace~\cite{berton2022rethinking}             & VGG16         & 1514 ms  & 64.7 & 83.4 \\
        CosPlace~\cite{berton2022rethinking}             & ResNet101     & 1488 ms  & 70.9 & 81.9 \\
        SALAD~\cite{Izquierdo_2024_salad}           & DINOv2        & 4805 ms  & 87.6 & 93.5 \\
        \rowcolor{gray!10}
        SALAD*                                      & DINOv2        & 4823 ms  & 88.0 & 94.5 \\
        BOQ~\cite{ali2024boq}                       & DINOv2        & 21333 ms & 83.7 & 92.8 \\
        \rowcolor{gray!10}
        BOQ*                                        & DINOv2        & 21047 ms & 88.8 & 93.7 \\
        \midrule
        \multicolumn{5}{l}{\textit{Mixed pipeline}} \\
        D\&C + CosPlace~\cite{Trivigno_2023_divideclassify} & EfficientNet & \textbf{30} ms & 71.4 & 87.6 \\
        DAPR - M                                    & EfficientNet  & 51 ms       & 74.5 & 88.1 \\
        DAPR - M                                    & ResNet101     & 54 ms       & 76.3 & 88.0 \\
        \rowcolor{gray!10}
        \textbf{DAPR - M}                           & DINOv2 & 74 ms & \textbf{89.7} & \textbf{94.3} \\
        \toprule
    \end{tabular}
    \vspace{0.5em}
    \caption{Results on SF-XL~\cite{berton2022rethinking} test v1 and v2 across three VPR types: classification-only (top), retrieval-only (middle), and mixed pipeline (bottom). SALAD$^*$ and BoQ$^*$ denote retraining with proposed LB loss $\mathcal{L}_{lb}$. \textbf{DAPR-M}: mixed pipeline with LB Loss and CFD.}
    \vspace{-5mm}
    \label{tab:performance_comparison}
\end{table}

\vspace{-2mm}
\subsection{Comparison with VPR State-of-the-arts}
\label{sec:comparison}
We compare the proposed DAPR framework against a broad set of VPR methods on the SF-XL benchmark. As indicated in Table~\ref{tab:performance_comparison}, DAPR-M (our mixed-pipeline framework) with DINOv2 achieves ~\textbf{89.7\%} R@1 on test v1 and \textbf{94.3\%} R@1 on test~v2, surpassing all compared methods that do not
incorporate distribution-aware training. 
Integrating $\mathcal{L}_{lb}$ as a plug-in to retrieval-only methods brings notable gains: SALAD$^*$ (SALAD~\cite{Izquierdo_2024_salad} trained with LB Loss) improves from 87.6\% to 88.0\% on test v1 and from 93.5\% to 94.5\% on test v2, while BoQ$^*$ (BoQ~\cite{ali2024boq} trained with LB Loss) improves from 83.7\% to 88.8\% and from 92.8\% to 93.7\%, respectively. These improvements confirm that long-tailed geographic imbalance is a shared bottleneck across VPR methods, not a weakness specific to any single model. DAPR-M further surpasses BoQ$^*$, demonstrating that classifier-based candidate filtering offers additional gains by narrowing the search space to geographically relevant candidates, enabling more precise retrieval than searching the full database.  

\begin{table}[h]
\centering
\scriptsize
\setlength{\tabcolsep}{3.5pt}
\renewcommand{\arraystretch}{1.3}
\begin{tabular}{l|cc|cc|cc}
\toprule
\rule{0pt}{2ex}
\multirow{2}{*}{\textbf{Method}} &
\multicolumn{2}{c|}{\textbf{R@1}} &
\multicolumn{2}{c|}{\textbf{R@5}} &
\multicolumn{2}{c}{\textbf{R@10}} \\
& \textbf{v1} & \textbf{v2}
& \textbf{v1} & \textbf{v2}
& \textbf{v1} & \textbf{v2} \\
\toprule
\rule{0pt}{2.3ex}
CosPlace
& 74.4 & 90.0 & 82.0 & 94.6 & 84.5 & 97.0 \\
\quad\textit{+ LB Loss}
& \textbf{76.1}~(+1.7) & 90.0
& \textbf{84.2}~(+2.2) & \textbf{95.8}~(+1.2)
& \textbf{86.3}~(+1.8) & 97.0 \\
\midrule
DC
& 44.1 & 64.0 & 57.2 & 84.9 & 63.6 & 90.1 \\
\quad\textit{+ LB Loss}
& \textbf{48.2}~(+4.1) & \textbf{66.2}~(+2.2)
& \textbf{64.2}~(+7.0) & \textbf{86.1}~(+1.2)
& \textbf{69.6}~(+6.0) & \textbf{90.3}~(+0.2) \\
\midrule
DC (mixed)
& 55.1 & 81.1 & 62.2 & 88.6 & 64.7 & 90.0 \\
\quad\textit{+ LB Loss}
& \textbf{57.4}~(+2.3) & \textbf{81.3}~(+0.2)
& \textbf{65.4}~(+3.2) & \textbf{89.0}~(+0.4)
& \textbf{66.9}~(+2.2) & \textbf{91.1}~(+1.1) \\
\bottomrule
\end{tabular}
\vspace{0.5em}
\caption{Generalization of LB loss across VPR frameworks on SF-XL test v1/v2.
Recall@N (\%) reported.}
\vspace{-4mm}
\label{tab:daloss_generalization}
\end{table}

\noindent\textbf{DAPR Across Pipelines and Backbones.}
Distribution-aware training provides consistent gains across all pipeline configurations.
In the classification-only setting, DAPR-C (our classification-only framework) with DINOv2 reaches 86.4\% R@1 on test~v1, outperforming D\&C~\cite{Trivigno_2023_divideclassify} (61.0\%) by +25.4\%. 
As shown in Table~\ref{tab:performance_comparison}, these improvements hold across EfficientNet, ResNet101, and DINOv2, indicating that $\mathcal{L}_{lb}$ addresses a data-level bottleneck largely orthogonal to backbone capacity.

\noindent\textbf{Memory and Inference-Time Analysis.}
In the mixed-pipeline on SF-XL, when DAPR-M is applied, the classifier filters the database to $\sim$450 candidates per query, requiring only \textbf{1.35\,MB} to store the 768-dim descriptors (FP32) for the following retrieval step. In contrast, retrieval-only methods must index the full database ($\sim$2.8M images): SALAD~\cite{Izquierdo_2024_salad} (8{,}448-dim) requires \textbf{94.8\,GB} and BoQ~\cite{ali2024boq} (8{,}192-dim) requires \textbf{91.9\,GB}, over four orders of magnitude larger due to both higher-dimensional descriptors (8K+ vs.\ 768 dimensions) and full-database indexing (2.8M vs.\ 450 images). 
The compact candidate pool also accelerates inference, as D\&C~\cite{Trivigno_2023_divideclassify} achieves 30\,ms and DAPR-M with DINOv2
achieves 74 ms per query, both over \textbf{$60\times$} faster than SALAD (4,805 ms), while delivering a 25.6\% relative R@1 improvement over the best mixed-pipeline baseline on test v1.

\noindent\textbf{Head-Tail Class Analysis.}
To pinpoint the source of improvement, we analyse performance across head, middle, and tail classes on SF-XL (Figure~\ref{fig:class_group_comparison}), using the same EfficientNet backbone in the mixed pipeline as D\&C for a fair comparison. DAPR delivers consistent gains across all three groups: +0.98\% R@1 on head, +1.30\% on middle, and +1.47\% on tail at R@1. This directly validates that our method targets the sparsely represented locations where existing models fail most. The gain pattern is more pronounced at R\@5, where tail classes improve by +7.35\% compared to +1.30\% for middle and +1.72\% for head.

\begin{table}[t]
    \centering
    \scriptsize
    \setlength{\tabcolsep}{2pt}
    \renewcommand{\arraystretch}{1.15}
    \begin{tabular}{l|c|c|ccc|ccc|ccc}
        \toprule
        \rule{0pt}{2.3ex}
        \multirow{2}{*}{\textbf{Method}} &
        \multirow{2}{*}{\textbf{Backbone}} &
        \multirow{2}{*}{\textbf{Dim.}} &
        \multicolumn{3}{c|}{\textbf{MSLS}} &
        \multicolumn{3}{c|}{\textbf{Pitts30k}} &
        \multicolumn{3}{c}{\textbf{Nordland}} \\
        & & &
        \textbf{R@1} & \textbf{R@5} & \textbf{R@10} &
        \textbf{R@1} & \textbf{R@5} & \textbf{R@10} &
        \textbf{R@1} & \textbf{R@5} & \textbf{R@10} \\
        \toprule
        CosPlace~\cite{berton2022rethinking}   & ResNet101 & 128
            & 81.7 & 89.7 & 91.6 & 86.7 & 94.5 & 97.0 & 41.1 & 56.3 & 63.3 \\
        \rowcolor{groupgray}
        CosPlace$^*$~\cite{berton2022rethinking} & ResNet101 & 128
            & \textbf{82.2} & \textbf{89.8} & \textbf{92.4}
            & \textbf{89.4} & \textbf{96.6} & \textbf{97.9}
            & \textbf{44.6} & \textbf{59.4} & \textbf{65.7} \\
        \midrule
        SALAD~\cite{Izquierdo_2024_salad}      & DINOv2 & 8448
            & 91.9 & 96.4 & 96.5 & 92.3 & 96.3 & 97.3 & 76.0 & 89.2 & 92.0 \\
        \rowcolor{groupgray}
        SALAD$^*$~\cite{Izquierdo_2024_salad}  & DINOv2 & 8448
            & \textbf{92.6} & \textbf{96.5} & \textbf{97.0}
            & \textbf{92.7} & \textbf{96.8} & \textbf{97.7}
            & \textbf{76.6} & \textbf{89.6} & \textbf{92.9} \\
        \midrule
        BoQ~\cite{ali2024boq}                  & DINOv2 & 8192
            & 91.2 & 95.3 & 96.1 & 92.6 & 96.4 & 97.4 & 81.3 & 92.5 & 94.8 \\
        \rowcolor{groupgray}
        BoQ$^*$~\cite{ali2024boq}              & DINOv2 & 8192
            & \textbf{93.7} & \textbf{96.2} & \textbf{96.8}
            & \textbf{92.9} & \textbf{96.6} & \textbf{97.5}
            & \textbf{83.7} & \textbf{94.6} & \textbf{96.5} \\
        \toprule
    \end{tabular}
    \vspace{0.5em}
    \caption{Comparison with state-of-the-art retrieval methods on MSLS,
    Pitts30k, and Nordland. $^*$ denotes models
    retrained with $\mathcal{L}_{lb}$. \textbf{Bold} indicates the better value
    within each method pair.}
    \label{tab:comparison_salad_boq}
    \vspace{-4mm}
\end{table}

\subsection{Generalization Across VPR Methods}
\label{sec:generalization}
To verify that the proposed DAPR framework addresses a fundamental problem rather than a method-specific weakness, we integrate the two plug-in modules independently into different VPR methods on both retrieval-only and mixed-pipeline settings.

\begin{figure}[tb]
  \centering
  \begin{minipage}{0.47\linewidth}
    \centering
    \includegraphics[width=\linewidth]{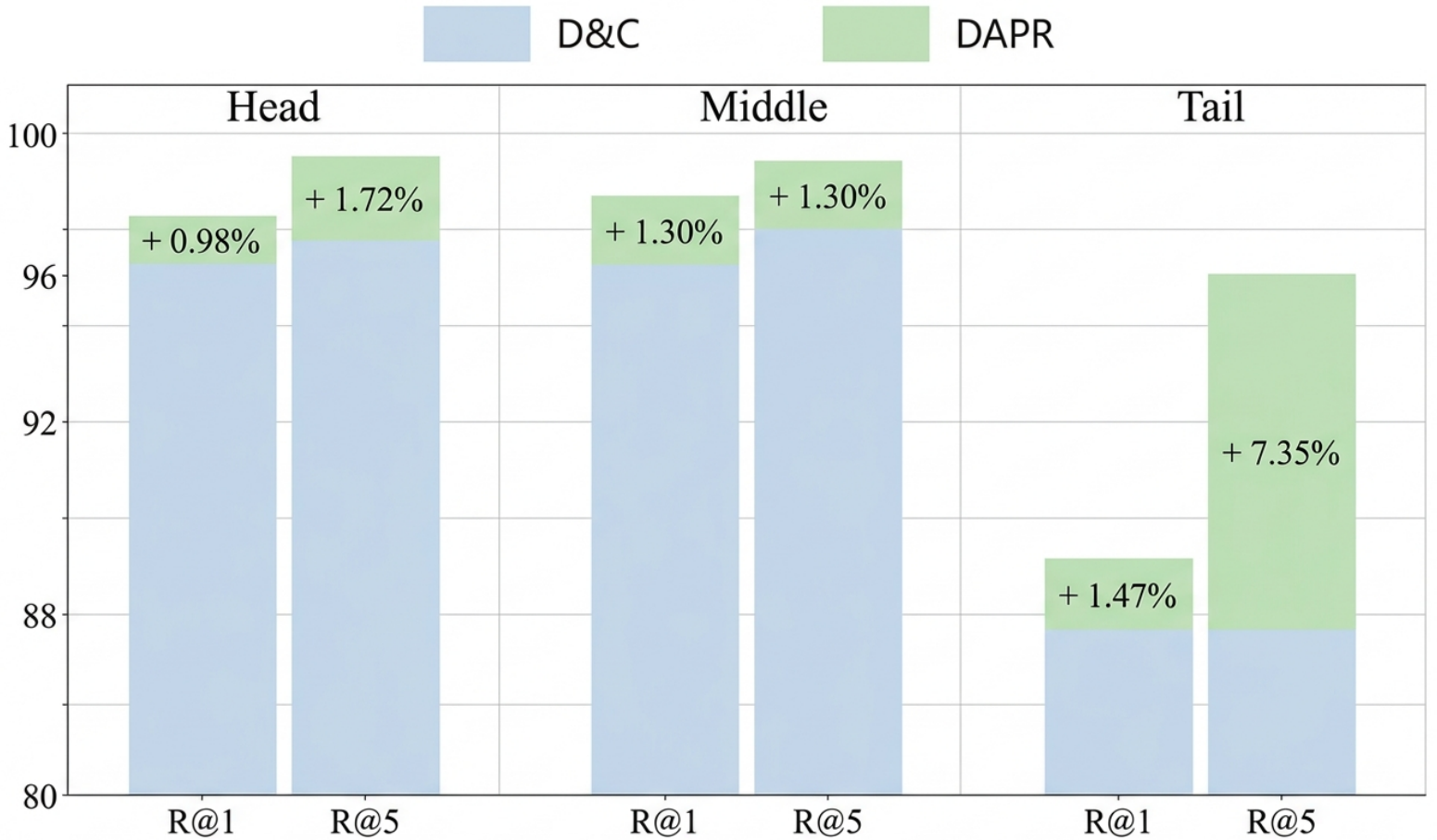}
    \caption{Performance comparison across head, middle, and tail classes on the SF\_XL dataset with Recall@N (\%).}
    \label{fig:class_group_comparison}
  \end{minipage}
  \hfill
  \begin{minipage}{0.45\linewidth}
    \centering
    \includegraphics[width=\linewidth]{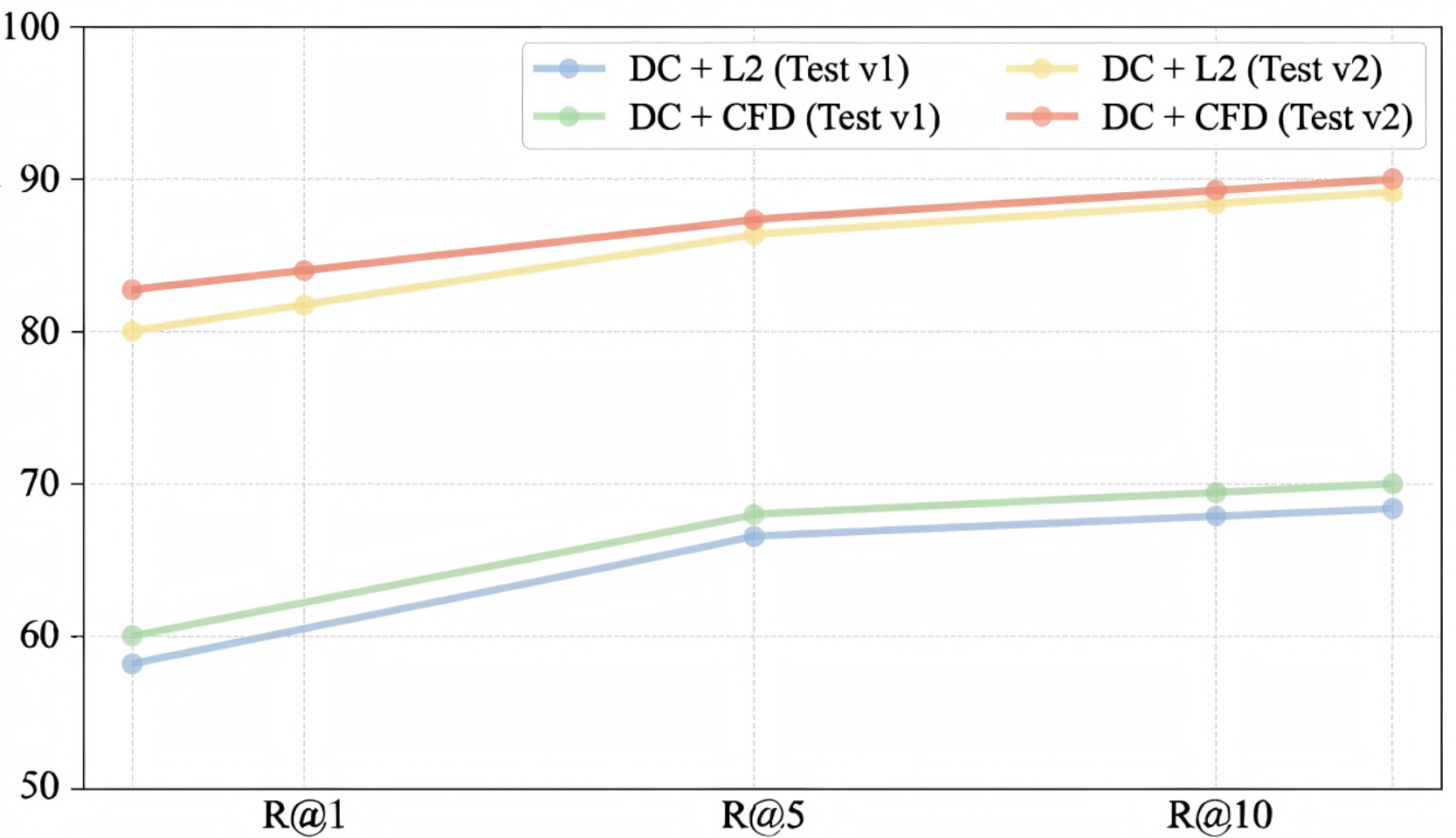}
    \caption{Comparison of Retrieval Distance on D\&C Method with SF-XL test v1 and test v2.}
    \label{fig:retrieval_distance}
  \end{minipage}
\vspace{-3mm}
\end{figure}

\noindent\textbf{Generalization on SF-XL.} 
Table~\ref{tab:daloss_generalization} presents results on the large-scale SF-XL~\cite{berton2022rethinking} benchmark. Applying $\mathcal{L}_{lb}$ to CosPlace brings a gain of +1.7\% R@1 on test~v1. On the D\&C baseline, the improvement is larger (+4.1\% R@1, +7.0\% R@5 on test~v1), as classification heads rely on per-class decision boundaries and are thus more sensitive to class imbalance than contrastive methods that inherently emphasize hard examples. Combining $\mathcal{L}_{lb}$ with the D\&C mixed pipeline further adds +2.3\% R@1, and gains are consistent across both test sets, confirming robustness to query-set variation. As shown in Figure~\ref{fig:retrieval_distance}, replacing L2 distance with CFD in both D\&C and our mixed pipeline achieves consistent gains across R\@1, R\@5,  on both mixed-pipeline methods (DAPR and D\&C), indicating that CFD provides reliable distribution-aware matching without any additional training or label supervision. 
More baseline results including CliqueMining~\cite{izquierdo2024close}, Megaloc~\cite{berton2025megaloc}, CricaVPR~\cite{lu2024cricavpr}, and ImAge~\cite{lu2026towards} are indicated in the supplementary materials.

\noindent \textbf{Generalization on Small-scale Benchmarks.} 
We integrate DAPR as a plug-in module into two state-of-the-art contrastive retrieval-only methods, SALAD~\cite{Izquierdo_2024_salad} and BoQ~\cite{ali2024boq}. Both are trained on GSV-Cities~\cite{ali2022gsv} and evaluated on MSLS~\cite{warburg2020mapillary} and Pitts30k~\cite{torii201524}. As indicated in Table~\ref{tab:comparison_salad_boq}, On MSLS, SALAD with $\mathcal{L}_{lb}$ (SALAD$^*$), BoQ with $\mathcal{L}_{lb}$ (BoQ$^*$) achieves +2.5\% R@1, and SALAD$^*$ gains +0.7\%. On Pitts30k, SALAD$^*$ and BoQ$^*$ improve by +0.4\% and +0.3\%, respectively. The improvements on both benchmarks, which cover diverse urban environments, confirm that $\mathcal{L}_{lb}$ reliably addresses geographic class imbalance as a lightweight, method-agnostic plug-in with different scales of datasets.

\noindent \textbf{Tail-Class Analysis on Small-scale Benchmarks.}
Table~\ref{tab:longtail} isolates gains on tail classes, the sparsely sampled zones identified in Figure~\ref{fig:class_spatial_distribution} as most safety-critical.
On MSLS, BoQ$^*$ improves tail-class R@1 by +2.70\% and SALAD$^*$ by +0.90\%.
On Pitts30k, BoQ$^*$ and SALAD$^*$ gain +3.37\% and +0.63\%, respectively.
The fact that tail-class gains consistently exceed overall gains confirms that $\mathcal{L}_{lb}$ primarily corrects the failure mode it was designed for, rather than providing uniform across-the-board improvements.

\begin{table}[tb]
\centering
\scriptsize
\setlength{\tabcolsep}{4pt}
\begin{tabular}{lcccccc}
\toprule
& \multicolumn{3}{c}{MSLS} & \multicolumn{3}{c}{Pitts30k} \\
\cmidrule(lr){2-4} \cmidrule(lr){5-7}
Method & R@1 & R@5 & R@10 & R@1 & R@5 & R@10 \\
\midrule
SALAD  & 86.49 & 93.24 & 94.14 & 90.56 & 94.86 & 96.87 \\
SALAD* & \textbf{87.39} & \textbf{94.14} & \textbf{95.05}
       & \textbf{91.19} & \textbf{95.50} & \textbf{97.36} \\
       & (+0.90) & (+0.90) & (+0.91) & (+0.63) & (+0.64) & (+0.49) \\
\midrule
BoQ    & 89.64 & 93.69 & 94.59 & 89.19 & 93.69 & 94.59  \\
BoQ*   & \textbf{91.44} & \textbf{94.59} & \textbf{95.50} 
       & \textbf{92.56} & \textbf{96.33} & \textbf{97.65}\\
       & (+1.80) & (+0.90) & (+0.91) & (+3.37) & (+2.64) & (+3.06)  \\
\bottomrule
\end{tabular}
\vspace{0.5em}
\caption{Tail-class performance on Pitts30k and MSLS. 
         SALAD* and BoQ* denote training with $\mathcal{L}_{lb}$.
         Parentheses show absolute gain over original methods.}
\label{tab:longtail}
\end{table}

\subsection{Ablation Study on DAPR}
\label{sec:ablation}
\noindent\textbf{Ablation study on individual modules.} We conduct comprehensive ablation studies to validate the individual contributions of the $\mathcal{L}_{lb}$ and the multi-scale distance retrieval module. Table~\ref{tab:Module_ablation} presents a cumulative study. Starting from a CrossEntropy baseline with $L_2$ retrieval (75.8\% R@1 on test~v1), $\mathcal{L}_{lb}$ alone yields +12.2\%, confirming that distribution-aware training is the primary driver of improvement. 
Replacing standard metrics with CFD yields performance gains in the proposed pipeline with a further +1.7\%. 

\begin{table}[b]
    \centering
    \scriptsize
    \setlength{\tabcolsep}{4pt}
    \renewcommand{\arraystretch}{1.15}
    \begin{tabular}{l|ccc|ccc}
        \toprule
        \rule{0pt}{2.3ex}
        \multirow{2}{*}{\textbf{VPR Method}} &
        \multicolumn{3}{c|}{\textbf{Test v1}} &
        \multicolumn{3}{c}{\textbf{Test v2}} \\
        & \textbf{R@1} & \textbf{R@5} & \textbf{R@10}
        & \textbf{R@1} & \textbf{R@5} & \textbf{R@10} \\
        \toprule
        \rule{0pt}{2.3ex}
        DINOv2 + CE
            & 75.8 & 83.4 & 85.6 & 91.1 & 93.8 & 95.7 \\
        \midrule
        \rule{0pt}{2.3ex}
        DINOv2 + LBLoss + L2
            & 88.0 & 89.1 & 90.6 & 93.8 & 96.2 & 97.0 \\
        \rowcolor{groupgray}
        DINOv2 + LBLoss + CFD
            & \textbf{89.7} & \textbf{92.4} & \textbf{92.5}
            & \textbf{94.3} & \textbf{97.0} & \textbf{97.5} \\
        \toprule
    \end{tabular}
    \vspace{0.5em}
    \caption{Ablation study on DAPR components. Baseline uses CrossEntropy
    loss with L2 retrieval. Results are reported as Recall@$N$ (\%).}
    \label{tab:Module_ablation}
    \vspace{2mm}
\end{table}

\noindent\textbf{Long-Tailed Learning methods Analysis.}
We compare $\mathcal{L}_{lb}$ against Cross-Entropy (CE), Logit Adjustment (LA)~\cite{laloss2020}, and Focal Loss (FL)~\cite{lin2017focal} under the full mixed pipeline, as indicated in Table~\ref{tab:loss_comparison}. $\mathcal{L}_{lb}$ outperforms CE by +13.9\% R@1 on test~v1 and +3.2\% on test~v2, and surpasses LA and FL by +1.7\% R@1 and +3.3\% R@1 on test~v1, respectively. The advantage over LA and FL, which each address only one failure mode (prediction bias or gradient imbalance), supports the design choice of coupling logit adjustment and sample reweighting through the shared class prior $\hat{p}_c$.
Sensitivity analyses for the hyperparameters of the two modules from DAPR, the Low-visit Bias Loss ($\mathcal{L}_{lb}$) and the CFD retrieval metric, are provided in the supplementary material.

\begin{table}[tb]
\centering
\scriptsize
\setlength{\tabcolsep}{2pt}
\renewcommand{\arraystretch}{1.3}
\begin{tabular}{l|cc|cc|cc}
\toprule
\rule{0pt}{2ex}
\multirow{2}{*}{\textbf{Loss Type}} &
  \multicolumn{2}{c|}{\textbf{R@1}} &
  \multicolumn{2}{c|}{\textbf{R@5}} &
  \multicolumn{2}{c}{\textbf{R@10}} \\
 & \textbf{v1} & \textbf{v2}
 & \textbf{v1} & \textbf{v2}
 & \textbf{v1} & \textbf{v2} \\
\toprule
CrossEntropy
  & 75.8 & 91.1 & 83.4 & 93.8 & 85.6 & 95.7 \\
LogitAdjust~\cite{laloss2020}
  & 88.0~(+12.2) & 93.3~(+2.2)
  & 91.5~(+8.1)  & \textbf{97.2}~(+3.4)
  & 92.2~(+6.6)  & \textbf{97.7}~(+2.0) \\
Focal Loss~\cite{lin2017focal}
  & 86.4~(+10.6) & 94.0~(+2.9)
  & 90.5~(+7.1)  & 96.3~(+2.5)
  & 91.4~(+5.8)  & 97.3~(+1.6) \\
\midrule
\rowcolor{gray!10}
\textbf{Ours (LB Loss)}
  & \textbf{89.7}~(+13.9) & \textbf{94.3}~(+3.2)
  & \textbf{92.4}~(+9.0)  & 97.0~(+3.2)
  & \textbf{92.5}~(+6.9)  & 97.5~(+1.8) \\
\bottomrule
\end{tabular}
\vspace{0.5em}
\caption{Comparison of LBLoss against CrossEntropy, LogitAdjust~\cite{laloss2020},
and Focal Loss~\cite{lin2017focal} on SF-XL test v1/v2. All methods use the same
DINOv2 backbone with our mixed pipeline. Parentheses
show absolute gain over the CrossEntropy baseline.}
\label{tab:loss_comparison}
\vspace{-4mm}
\end{table}

\begin{figure}[tb]
  \centering
  \includegraphics[width=0.5\linewidth]{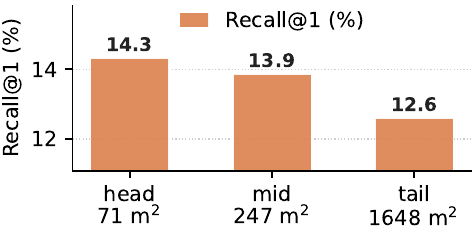}
  \caption{Per-class Recall@1 across head, middle, and tail bins defined by leaf spatial extent (m$^2$) on CPlaNet's~\cite{seo2018cplanet} of SF-XL (131{,}080 leaves).}
  \label{fig:partition}
\end{figure}

\noindent\textbf{Adaptive Partitioning vs $\mathcal{L}_{lb}$.}
Adaptive partitioning balances geographic classes by image count instead of at the loss level. We apply CPlaNet's~\cite{seo2018cplanet} kd-tree partition to SF-XL, it evens the per-class counts but
reshapes the imbalance into leaf area. As shown in Fig.~\ref{fig:partition}, Per-class Recall@1 drops from 14.3 to 12.6. Adaptive partitioning rebalances the data layout, whereas $\mathcal{L}_{lb}$ rebalances the learning signal, so the two address different scopes of imbalance.

\noindent\textbf{DAPR Classifier with SOTA Retrieval Head.} We keep the DAPR-C classifier to filter the top classes from the full database, then re-rank the shortlist with SALAD$^{*}$ or BoQ$^{*}$. Both bring only a marginal R@1 gain over DAPR-M, while the inference time grows more than 10 times due to their feature size requirements. Moreover, SALAD and BoQ reshape DINOv2 features into cluster aggregators that the DAPR classifier cannot reuse, so the pipeline maintains two independent DINOv2 backbones at both the training and inference stages. DAPR-M reaches comparable accuracy with a single backbone at far lower cost.

\vspace{-0.5em}
\section{Conclusion}
\vspace{-1em}

In this paper, we first systematically reveal and address the fundamental long-tailed challenge arising from natural geographic distributions in the VPR task. We propose DAPR framework including two plug-and-play modules: 1) Low-visit Bias Loss rebalances gradient contributions; 2) CFD for multi-scale distance search in mixed pipeline.
Extensive experiments on SF-XL, MSLS and Pitts30k demonstrate 
that DAPR can be independently integrated into existing VPR frameworks and generalizes across diverse backbone architectures, validating its robustness and broad applicability. 
Nevertheless, our current formulation defines the long-tail distribution based on \emph{sample count per geographic class}, focusing on dataset-level imbalance. Another direction worth exploring is the \emph{feature-level} long-tail phenomenon, where class difficulty is characterized by intra-class feature coherence rather than sample frequency.


\section*{Acknowledgments}
We acknowledge Lambda for providing us with compute for this work. 

\newpage
\bibliographystyle{splncs04}
\bibliography{main}









\clearpage
\begin{center}
{\large\bfseries Lost in the Tail: Addressing Geographic Imbalance in Urban Visual Place Recognition\par}
\vspace{0.4em}
{\Large\bfseries Supplementary Material\par}
\end{center}
\vspace{1.0em}
\suppressfloats[t]

\setcounter{page}{1}
\setcounter{section}{0}
\renewcommand{\thesection}{\Alph{section}}
\renewcommand{\thesubsection}{\Alph{section}.\arabic{subsection}}
\renewcommand{\thetable}{\Alph{section}\arabic{table}}
\renewcommand{\thefigure}{\Alph{section}\arabic{figure}}

\noindent The supplemental material provides additional details and experiments to complement the main manuscript. Section~\ref{sec:long-tail} analyzes the geographic class distribution in MSLS and GSV-Cities training sets beyond SF-XL.
Section~\ref{sec:quantitative} presents extended quantitative experiments, including sensitivity analyses for the key hyperparameters $\beta$ and $\kappa$ in $\mathcal{L}_{lb}$ and additional ablation analysis.
Section~\ref{sec:qualitative} provides qualitative analysis
through t-SNE feature space visualizations, representative retrieval examples on tail-class queries, and failure case studies.
Section~\ref{sec:setup} describes the full training configuration and hyperparameter settings corresponding to Section~4.1 of the main manuscript.
Section~\ref{sec:cfd-supp} details the stratified multi-scale frequency sampling strategy used in CFD during inference.

\section{Additional Long-tailed data distribution in VPR Datasets}
\label{sec:long-tail}
\vspace{-3mm}

\begin{figure}[t]
  \begin{subfigure}{0.49\linewidth}
    \includegraphics[width=\linewidth]{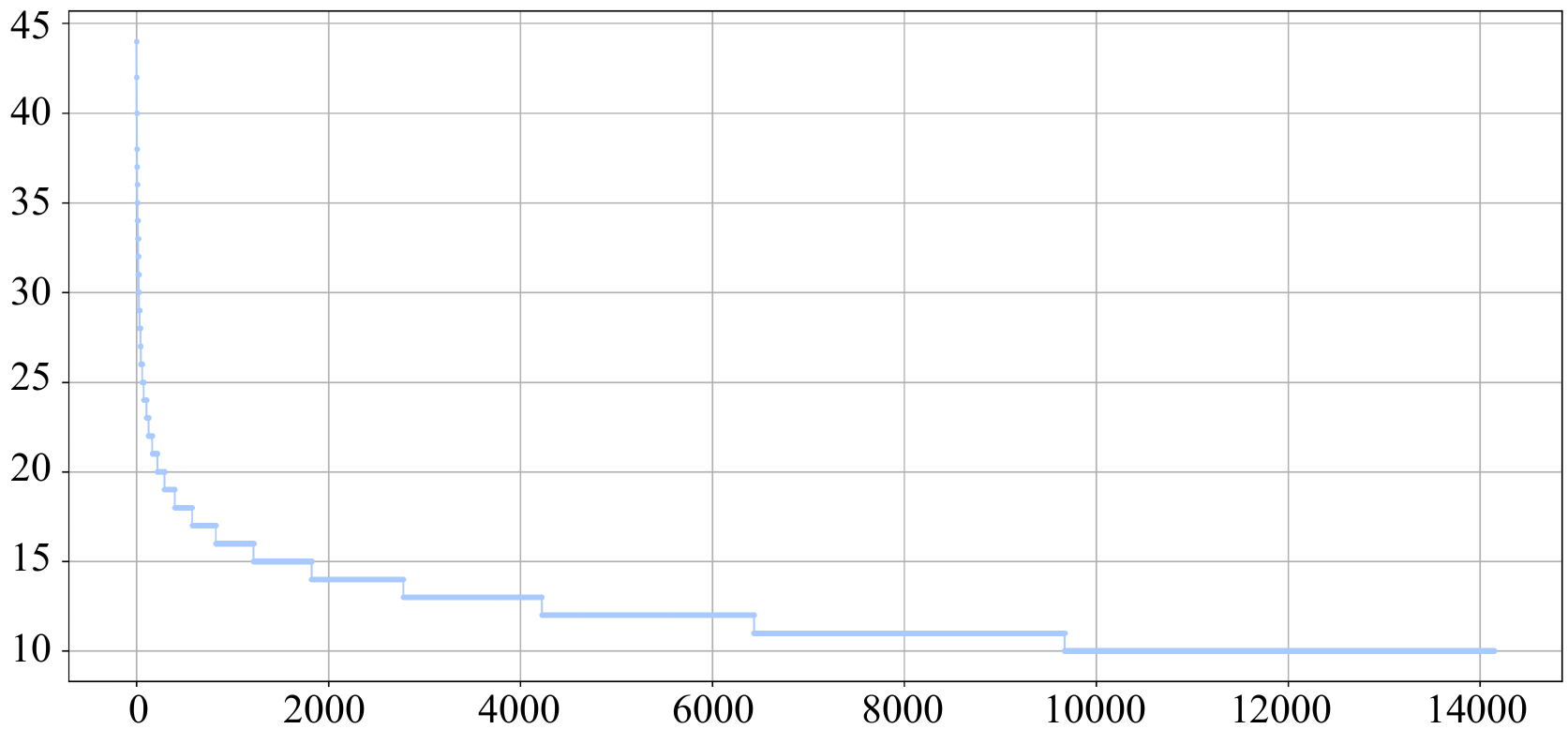}
    \caption{GSV Train Dataset Distribution}
  \end{subfigure}
  \hfill
  \begin{subfigure}{0.49\linewidth}
    \includegraphics[width=\linewidth]{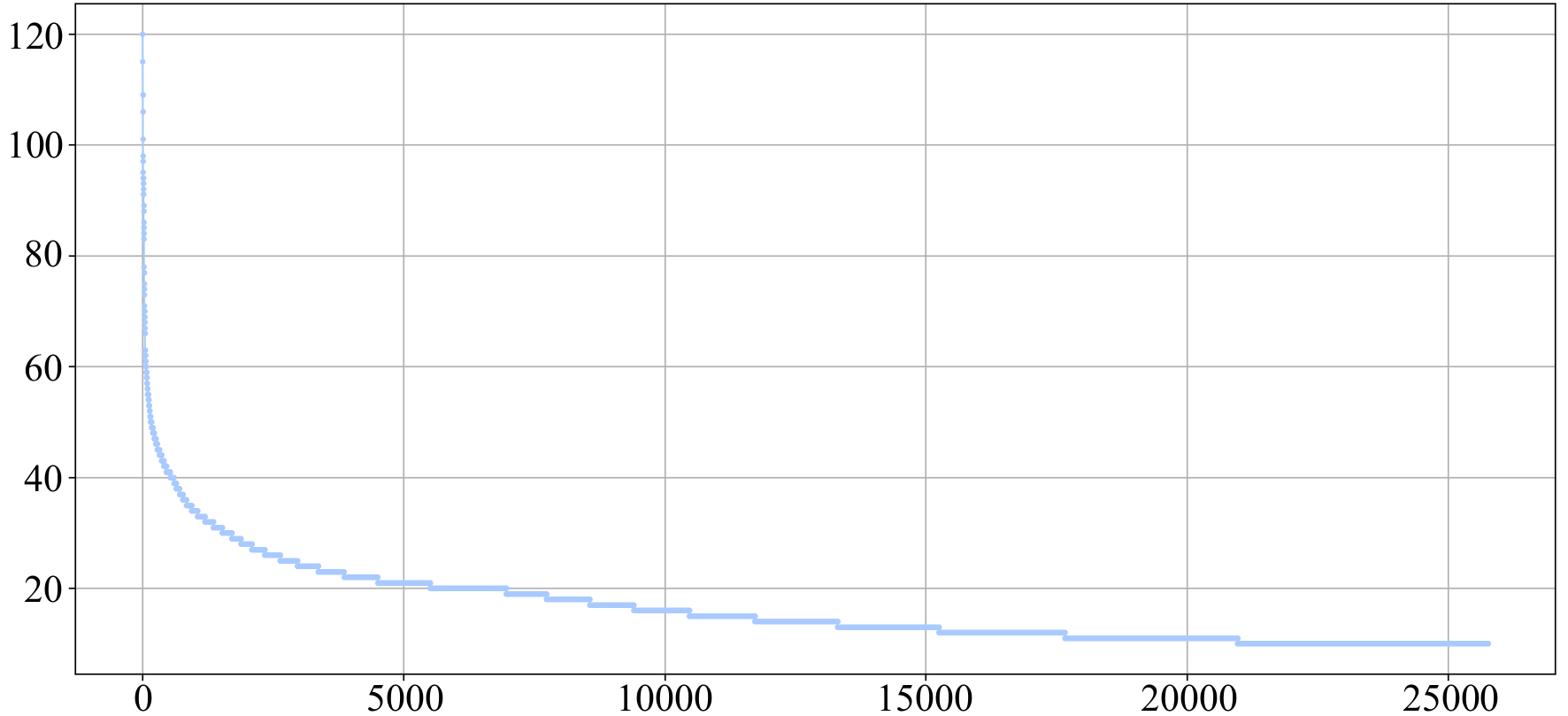}
    \caption{MSLS Train Dataset Distribution}
  \end{subfigure}
  \caption{Geographic class distribution in additional VPR benchmarks. (a) GSV-Cities training set and (b) MSLS training set. The x-axis represents class index sorted by frequency, and the y-axis shows the number of training images per class.}
  \label{fig:dataset_distribution}
\end{figure}

Complementing the SF-XL long-tail analysis in Section~1 and Figure~1 of the main manuscript, we analyze two widely-used visual place recognition benchmarks: MSLS-train~\cite{warburg2020mapillary} and GSV-Cities~\cite{ali2022gsv}. As shown in Figure~\ref{fig:dataset_distribution}, both datasets exhibit pronounced long-tailed distribution patterns similar to SF-XL, confirming that geographic class imbalance is an inherent characteristic of real-world location datasets. However, SF-XL differs fundamentally in its design philosophy and practical applicability. While MSLS and GSV-Cities are collected for small-scale, city-to-city retrieval across multiple metropolitan areas~\cite{warburg2020mapillary,ali2022gsv}, SF-XL provides large-scale, fine-grained localization within a single city using dense, unified geographic coverage with 20m$\times$20m cells~\cite{berton2022rethinking}. This design reflects real-world scenarios when building a VPR system for a large metropolitan area.

\begin{figure}[t]
 \includegraphics[width=\linewidth]{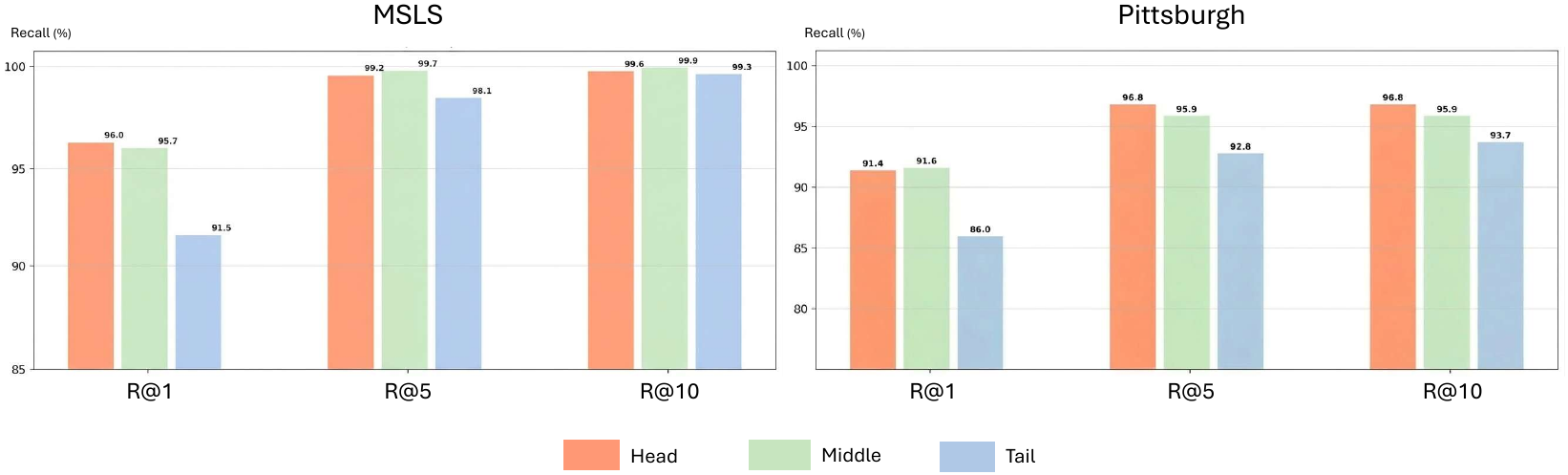}
  \caption{Head, middle, and tail class performance on MSLS and Pitts30k trained with SALAD~\cite{Izquierdo_2024_salad}.}
  \label{fig:longtail_msls_pitts}
\end{figure}

Despite these differences in dataset scale, the long-tailed structure in both MSLS and GSV-Cities is qualitatively consistent with SF-XL. Consistent with the head-tail analysis in Section~4.2 of the main manuscript, figure~\ref{fig:longtail_msls_pitts} presents a head-middle-tail breakdown of retrieval performance for SALAD on MSLS and Pitts30k. Across both benchmarks, the SALAD baseline exhibits a consistent performance gap across tiers, with tail classes lagging substantially at all recall cutoffs. As indicated in manuscript Tables~3 and~4 in the main manuscript, applying $\mathcal{L}_{lb}$ achieves tail-class gains that consistently exceed overall improvements.

\section{Additional Experiment Analysis}
\label{sec:quantitative}

We present detailed ablation studies on DAPR hyperparameters, additional sensitivity experiments of individual $\mathcal{L}_{lb}$ components, and tail-class analysis of the Multi-scale Distance retrieval module to further support our findings.

\subsection{Hyperparameter Sensitivity of $\mathcal{L}_{lb}$}

\begin{figure}[t]
    \includegraphics[width=\linewidth]{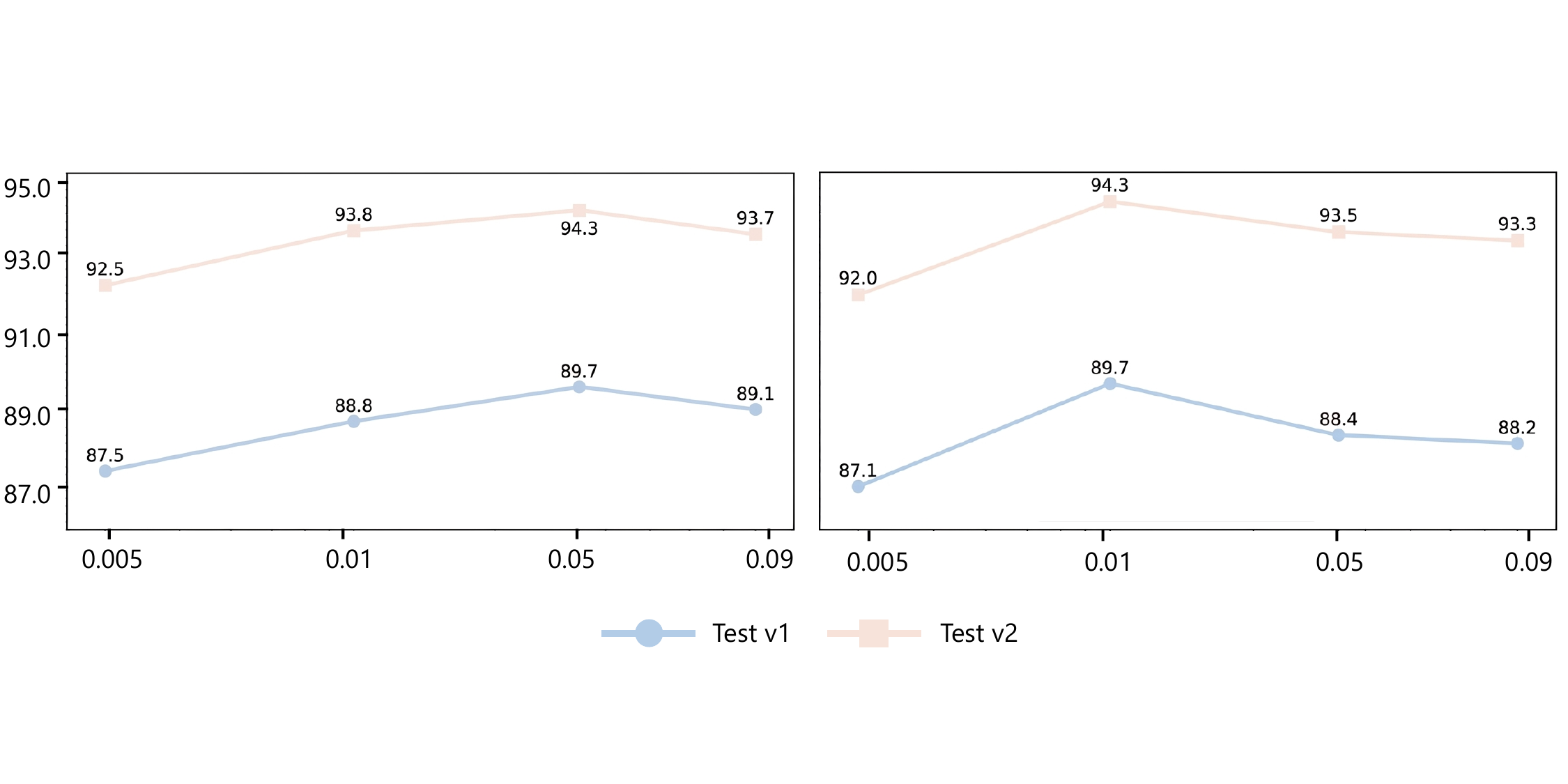}
  \vspace{-4em}
  \caption{Sensitivity analysis of $\mathcal{L}_{lb}$ hyperparameters on SF-XL. Left: varying the class reweighting exponent $\beta$ (Eq.~2). Right: varying the logit adjustment strength $\kappa$ (Eq.~3). R@1 (\%) remains stable across a wide range for both parameters.}
  \label{fig:loss_hparams}
\end{figure}

To complement Section~4.4 of the main manuscript and characterize the sensitivity of $\mathcal{L}_{lb}$ to its two key hyperparameters, we conduct ablation studies on the reweighting strength $\beta$ (Eq.~2 of the main manuscript Section~3.2), which controls the degree to which tail classes receive amplified gradient contributions, and the logit adjustment strength $\kappa$ (Eq.~3 of the main manuscript Section~3.2), which corrects classifier bias introduced by class prior imbalance. As shown in Figure~\ref{fig:loss_hparams}, performance on test v1 peaks at $\beta = 0.01$ and decreases gradually toward $\beta = 0.09$; test v2 follows a similar but less pronounced trend. This demonstrates that moderate reweighting suffices to balance gradient contributions across head and tail classes without over-suppressing frequent ones.

\subsection{Generalization to Recent Baselines and Nordland}
\begin{figure}[t]
  \centering
  \begin{subfigure}{0.32\linewidth}
    \includegraphics[width=\linewidth]{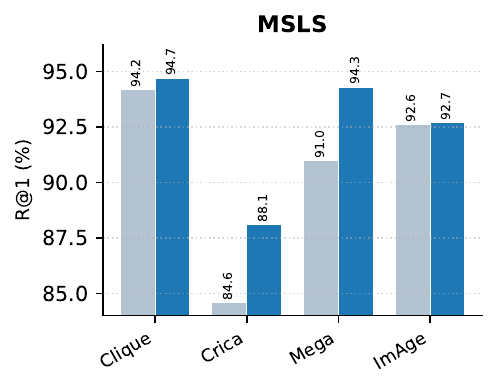}
    \caption{MSLS}
  \end{subfigure}\hfill
  \begin{subfigure}{0.32\linewidth}
    \includegraphics[width=\linewidth]{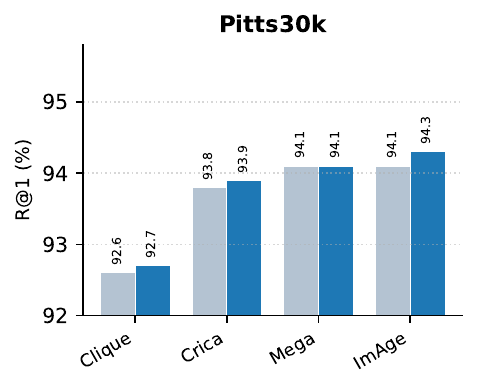}
    \caption{Pitts30k}
  \end{subfigure}\hfill
  \begin{subfigure}{0.32\linewidth}
    \includegraphics[width=\linewidth]{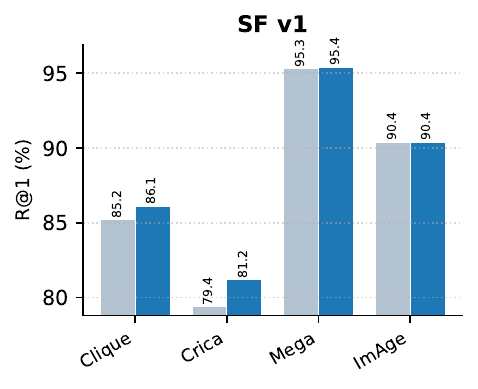}
    \caption{SF-XL v1}
  \end{subfigure}

  \vspace{0.8em}
  \begin{subfigure}{0.32\linewidth}
    \includegraphics[width=\linewidth]{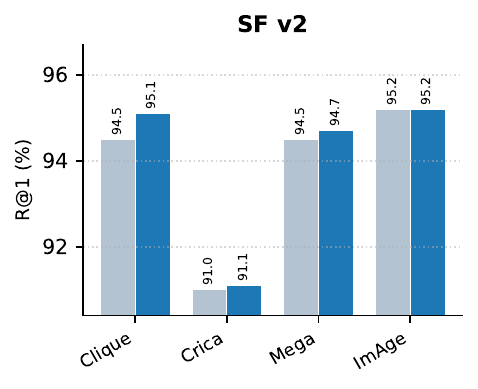}
    \caption{SF-XL v2}
  \end{subfigure}\hspace{0.03\linewidth}
  \begin{subfigure}{0.32\linewidth}
    \includegraphics[width=\linewidth]{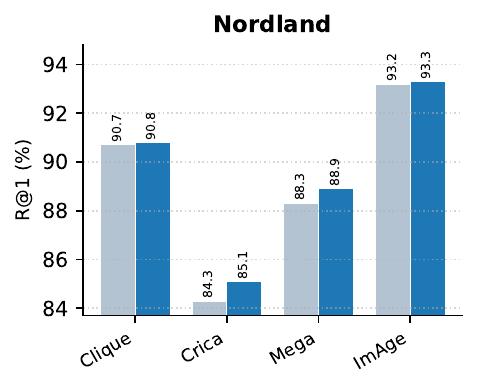}
    \caption{Nordland}
  \end{subfigure}
  \caption{R@1 of four recent VPR baselines (CliqueMining, CricaVPR, MegaLoc,
  ImAge) with and without the LB loss across five benchmarks.}
  \label{fig:baseline_lb_grid}
\end{figure}

We plug the LB loss into four recent VPR baselines and evaluate on five
benchmarks. As
Figure~\ref{fig:baseline_lb_grid} shows, the LB loss lifts or matches R@1 on SOTA baselines, which confirms that it transfers as a method-agnostic plug-in method.

\subsection{Additional Sensitivity Experiments}

\begin{figure}[t]
\centering
    \includegraphics[width=0.9\linewidth]{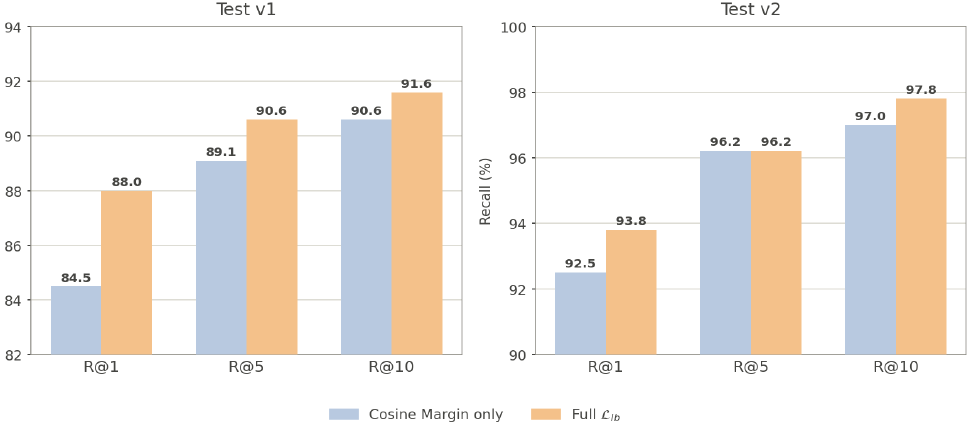}
  \caption{Ablation of the Cosine Margin component within $\mathcal{L}_{lb}^{cls}$
on SF-XL test v1/v2 within original L2 search. All results are Recall@N (\%).}
  \label{fig:ablation_cosine_margin}
\end{figure}

To isolate the contribution of the reweighting and logit adjustment terms in $\mathcal{L}_{lb}^{cls}$ (Eq.~6, Section~3.2), we compare a cosine-margin-only variant against the full $\mathcal{L}_{lb}$ on the mixed pipeline; results are shown in Figure~\ref{fig:ablation_cosine_margin}. Full $\mathcal{L}_{lb}$ achieves consistent gains across all recall cutoffs on both test sets. This confirms that inverse-frequency reweighting and prior-based logit adjustment provide gains orthogonal to cosine margin, and that all three components are necessary for the full benefit of $\mathcal{L}_{lb}$.

\subsection{CFD Tail-Class Retrieval Advantage.}

CFD reads each candidate through the mean characteristic-function amplitude $\bar{A}_j$ of its descriptor (Section~3.3). Head classes form dense clusters with a high and stable $\bar{A}_j$, whereas dispersed tail classes give a lower and more variable $\bar{A}_j$. The adaptive weight $\alpha_w^{(j)}=\min(\alpha\,r^{(j)},1)$ with $r^{(j)}=\bar{A}_q/\bar{A}_j$ saturates on tail candidates, where $\bar{A}_j$ is small, so the CFD distance is dominated by the amplitude term that captures cluster compactness. This explains why CFD improves tail-class retrieval.

\begin{figure}[t]
    \centering
    \includegraphics[width=0.75\linewidth]{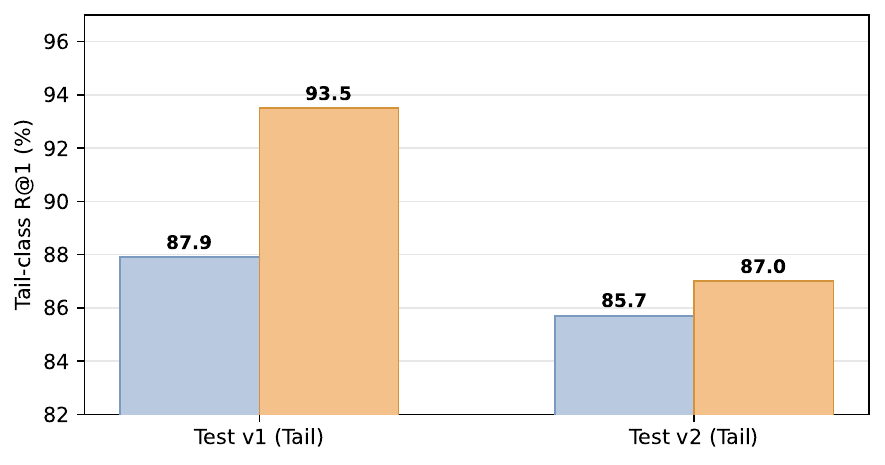}
    \vspace{0.5em}
    \caption{Comparison of tail-class R@1 (\%) between L2 and CFD retrieval on SF-XL test v1 and v2. CFD provides consistent gains on
    tail classes across both query sets.}
    \label{fig:cfd_tail_bar}
\end{figure}

As shown in Fig.~\ref{fig:cfd_tail_bar}, this advantage holds on sparsely sampled locations, complementing the overall CFD-over-L2 gain reported in Table~5 and Figure~5 of the main manuscript. On SF-XL test v1, CFD achieves 93.5\% tail-class R@1 compared to 87.9\% for L2 (+5.6\%), and on test v2, 87.0\% versus 85.7\% (+1.3\%). The consistently larger gain on v1 reflects that multi-scale distance estimation is particularly effective when tail-class features are geometrically dispersed. For the base amplitude weight $\alpha$, performance remains stable across $\alpha \in [0.3, 0.7]$, indicating our metric is effective on the large-scale VPR task.

\section{Qualitative Analysis}
\label{sec:qualitative}
We conduct comprehensive qualitative analysis through visualizations and case studies focusing on tail class behavior. Our analysis includes three aspects: feature space visualization via t-SNE, successful retrieval examples on underrepresented locations, and failure case analysis identifying remaining challenges.

\begin{figure}[t]
  \begin{subfigure}{0.48\linewidth}
    \includegraphics[width=\linewidth]{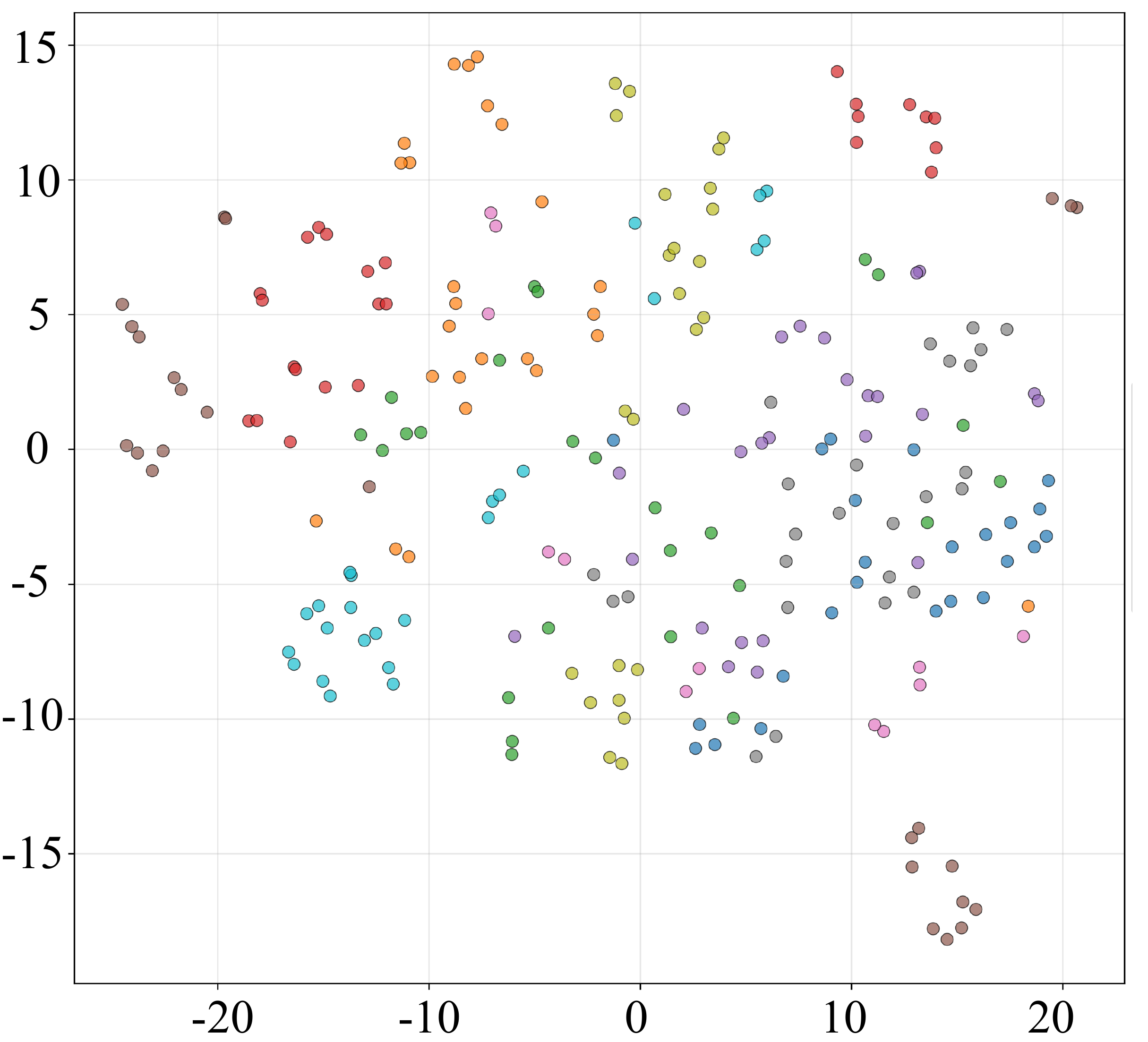}
    \caption{Baseline exhibits scattered features with substantial inter-class overlap.}
    \label{fig:tsne_inception}
  \end{subfigure}
  \hspace{0.03\linewidth}
  \begin{subfigure}{0.46\linewidth}
    \includegraphics[width=\linewidth]{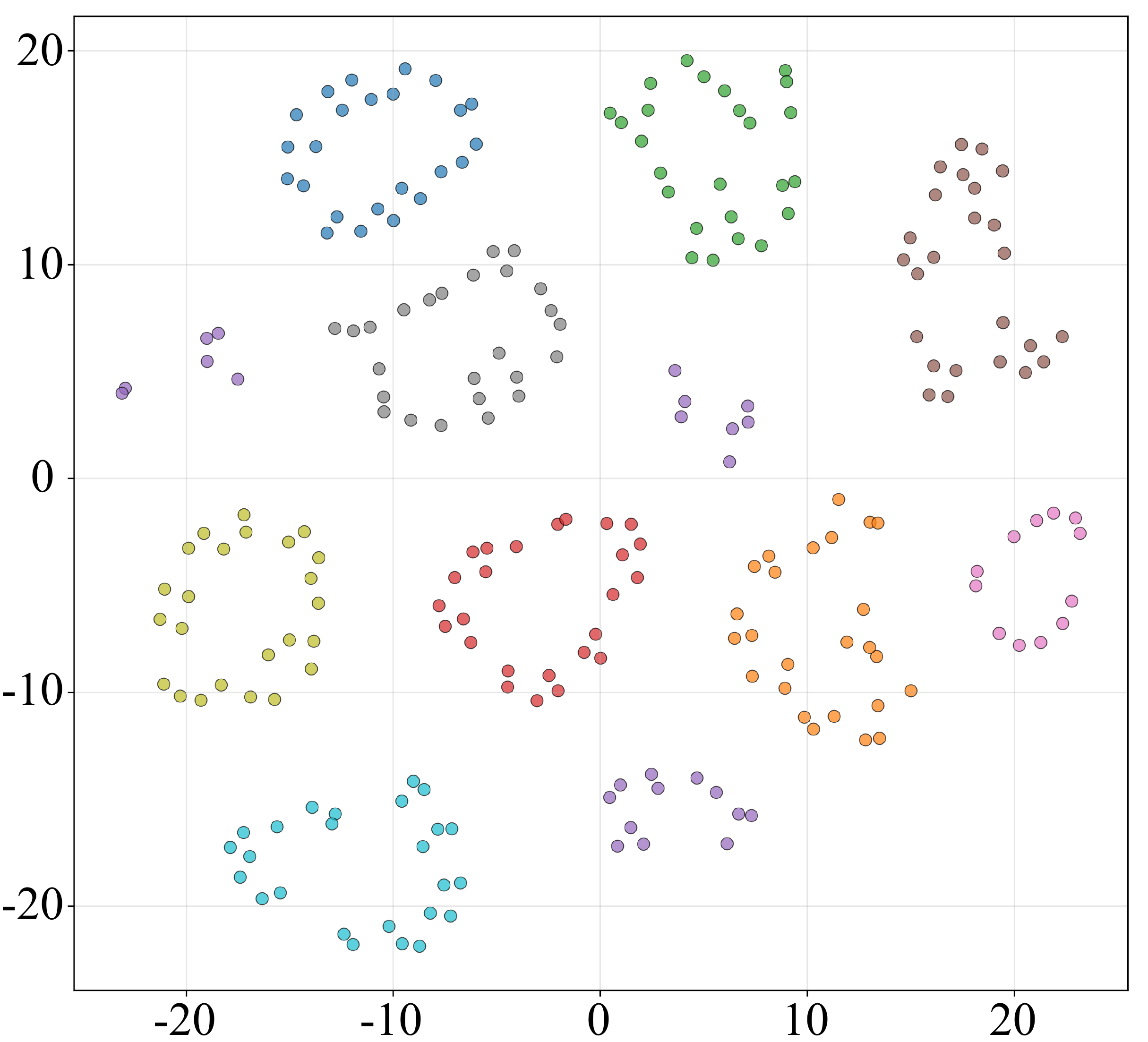}
    \caption{With Distribution-Aware Loss: The same tail classes form compact, well-separated clusters.}
    \label{fig:tsne_dino}
  \end{subfigure}
  \caption{t-SNE visualization of feature space organization for ten representative tail classes. The baseline (a) produces scattered representations with significant overlap between classes, while our distribution-aware approach (b) yields distinct, compact clusters. Each color represents one tail class identified by its UTM coordinate center [Easting, Northing] in meters. Classes shown: [544320, 4176980], [545960, 4175880], [550320, 4174860], [547420, 4179320], [552520, 4174960], [546800, 4183940], [554720, 4176080], [546020, 4176840], [548380, 4179960], [555720, 4175740].}
  \label{fig:tsne}
\end{figure}

\noindent\textbf{Feature Space Visualization.}
Figure~\ref{fig:tsne} shows t-SNE visualizations of feature embeddings for tail classes. We randomly sampled tail classes from the training set. Specific class IDs are shown in the caption.

Before applying distribution-aware learning (Figure~\ref{fig:tsne}a), tail class features exhibit significant scatter and inter-class overlap, indicating poor discriminability due to insufficient gradient contributions. After incorporating DAPR (Figure~\ref{fig:tsne}b), tail class features form more compact and well-separated clusters, demonstrating that DAPR framework successfully amplifies learning signals from rare geographic locations and produces more discriminative representations

\begin{figure}[t]
    \centering
    \includegraphics[width=0.9\linewidth]{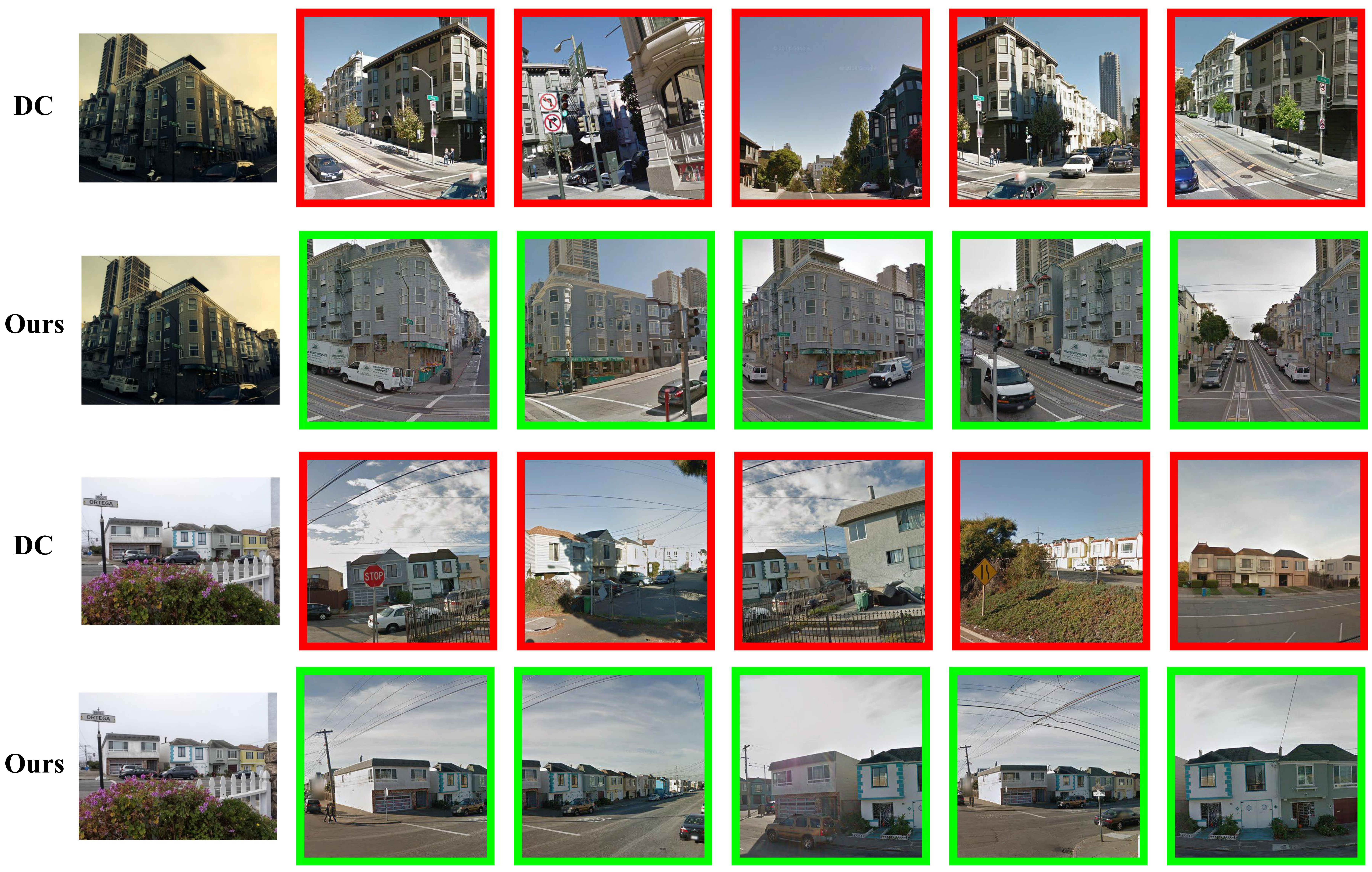}
    \caption{Qualitative comparison on tail class queries demonstrating DAPR's superiority over Divide\&Classify (D\%C) baseline. Each example shows: (left) query image from a tail class, (top row) D\&C's top-5 retrievals, (bottom row) DAPR's top-5 retrievals. Green borders indicate correct matches within 25m threshold; red borders denote failures.}
    \label{fig:good_cases}
\end{figure}

\noindent \textbf{Visual and Qualitative Evaluation.}
We provide qualitative examples specifically for tail class queries to illustrate DAPR's effectiveness on underrepresented locations. Figure~\ref{fig:good_cases} presents two representative tail class queries where DAPR substantially outperforms the D\&C baseline~\cite{Trivigno_2023_divideclassify}. For each query (leftmost column), we show the top-5 retrievals from both D\&C (top row) and DAPR (bottom row), with green borders indicating successful matches within the 25-meter threshold and red borders denoting failures. In the first example, D\&C retrieves five incorrect locations from visually similar but geographically distant areas, whereas DAPR successfully identifies five correct matches of the same apartment building from different viewpoints. These cases demonstrate that DAPR effectively mitigates the bias toward overrepresented classes.

\begin{figure}[t]
    \centering
    \includegraphics[width=0.9\linewidth]{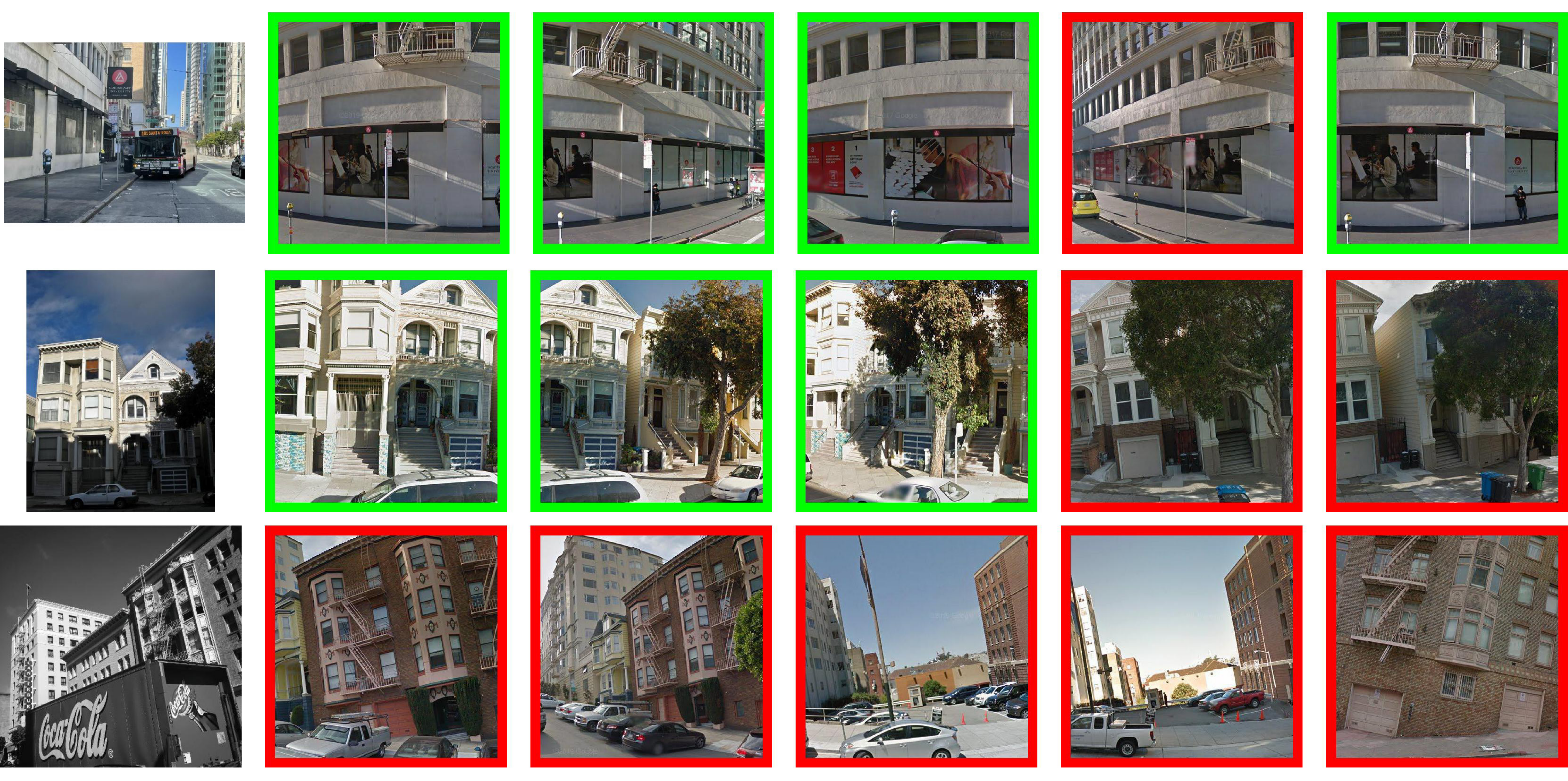}
    \caption{Failure cases revealing DAPR's limitations under extreme appearance variations. Tail class queries (left) with DAPR's top-5 retrievals (right).}
    \label{fig:bad_cases}
\end{figure}

\noindent \textbf{Failure Mode Analysis and Limitations.}
Despite significant improvements, Figure~\ref{fig:bad_cases} reveals three challenging scenarios where appearance variation remains problematic. The first query (top row) is a near-miss: DAPR retrieves four correct matches, and the fifth identifies the correct building facade but falls just outside the 25-meter evaluation threshold, so this case is largely an evaluation-radius artifact rather than a genuine retrieval error. The second (middle row) and third (bottom row) queries are true failures, caused respectively by dense foliage occlusion and by extreme color and illumination shifts between query and database. Both reflect a mismatch in visual appearance between the query and its database images. Closing this gap is the aim of the feature-level long-tail direction we outline in the conclusion, where a class is hard because its images look inconsistent rather than because it is rarely photographed.

\section{Experiment Setup Details}
\label{sec:setup}
\vspace{-3mm}
To complement the Section 4.1 and 4.3. We fine-tune the last few blocks of the backbone together with a pooling layer and a classification head, following the design of
CosPlace~\cite{berton2022rethinking} and D\&C~\cite{Trivigno_2023_divideclassify}.
At the classification step, the classifier prunes unrelated classes, and the fine-tuned backbone embedding directly forms the descriptor $f$ used for retrieval.
During training, we apply data augmentation including random horizontal flipping, color jittering, and random cropping.
For the classification-based pipeline, the rebalancing strength
$\beta$ and logit bias strength $\kappa$ in $\mathcal{L}_{lb}^{cls}$ are set to $0.01$ and $0.05$, respectively. To integrate $\mathcal{L}_{lb}^{rt}$
into retrieval-only methods SALAD~\cite{Izquierdo_2024_salad} and BoQ~\cite{ali2024boq}, we retain their original architectures and update the anchor weight $w_{y_i}$ and negative similarity bias $\nu_{y_j}$ at each epoch following Eq.~(2) and Eq.~(3) in the main manuscript, with bias strength set to $\kappa = 0.01$. During retrieval, the weights for our multi-scale adaptive distance are set to $\alpha = 0.7$ (amplitude) and $\lambda = 0.3$ (phase).

\section{Details of CFD Sampling and Uncertainty Estimation}
\label{sec:cfd-supp}

This section provides more details of the CFD strategy including sampling, which is included in Section 3.3 of the main manuscript.

To capture feature characteristics across multiple scales, we construct a frequency set $\mathbf{T} = \{\mathbf{t}_k\}_{k=1}^K$ through stratified sampling, where $t_k$ denotes the frequency vectors. We generate $K$ frequency samples from a mixture of Gaussian distributions:
\begin{equation}
\mathbf{t}_k \sim \mathcal{N}\left(\mathbf{0}, \sigma_0^2\mathcal{I}_d\right), \quad k = 1, \ldots, K
\end{equation}
where $\mathcal{I}_d \in \mathbb{R}^{d \times d}$ is the identity matrix and $\sigma_0 = \pi/4$ is the base scale parameter.

To capture patterns at different spatial granularities, we augment the base samples with frequency points at four logarithmically-spaced scales. The scales are defined as:
\begin{equation}
s_i = 10^{i-3}, \quad i \in \{1, 2, 3, 4\}
\end{equation}
\noindent yielding $\{10^{-2}, 10^{-1}, 10^0, 10^1\}$. For each scale $s_i$, we sample $K/4$ additional frequency points:
\begin{equation}
\mathbf{t}_k^{(s_i)} \sim \mathcal{N}(\mathbf{0}, (s_i\sigma_0)^2\mathcal{I}_d), \quad k = 1, \ldots, K/4
\end{equation}

This produces a candidate pool of $K + 4 \times (K/4) = 2K$ samples. All samples are L2-normalized: $\mathbf{t}_k \leftarrow \mathbf{t}_k / \|\mathbf{t}_k\|_2$. The final frequency set $\mathbf{T}$ consists of the first $K$ normalized samples from this pool, ensuring diverse frequency coverage while maintaining the desired cardinality. This sampling strategy enables small-scale frequencies ($s_1, s_2$) to capture fine-grained local patterns, while large-scale frequencies ($s_3, s_4$) encode global structure, which improves robust matching where tail classes have limited training samples.

\begin{table}[h]
\centering
\setlength{\tabcolsep}{8pt}
\label{tab:cfd_uncertainty}
\begin{tabular}{l|ccc}
\toprule
AUC-PR & $L_2$ & SUE & CFD\\
\midrule
Nordland & 0.80 & \textbf{0.86} & \textbf{0.86} \\
Pitts30k & 0.97 & 0.97          & \textbf{0.99} \\
MSLS     & 0.95 & 0.95          & \textbf{0.97} \\
\bottomrule
\end{tabular}
\caption{Image-matching uncertainty (AUC-PR, higher is better) of $L_2$, SUE, and CFD as inference-time confidence signals on three VPR datasets with a DINOv2 backbone.}
\end{table}

\noindent\textbf{CFD as an inference confidence signal.}
We report the AUC-PR score of CFD, SUE and the $L_2$ baseline on three VPR datasets using a DINOv2 backbone. As shown in the above table, CFD matches or surpasses SUE and $L_2$, confirming that CFD is a valid inference confidence signal.


\end{document}